\documentclass[opre]{informs3}
\usepackage{epsf}
\usepackage{epstopdf}
\usepackage{amsmath}
\usepackage{amssymb}
\usepackage{amsxtra}
\usepackage{algorithmic}
\usepackage{algorithm}
\usepackage{mathrsfs}
\usepackage{amsfonts}
\usepackage{dsfont}

\usepackage{subfigure}
\usepackage{latexsym}
\usepackage{float}
\usepackage{flushend}
\usepackage{cuted}
\usepackage{comment}
\usepackage{endnotes}
\let\footnote=\endnote
\let\enotesize=\normalsize
\def\notesname{Endnotes}%
\def\makeenmark{$^{\theenmark}$}
\def\enoteformat{\rightskip0pt\leftskip0pt\parindent=1.75em
  \leavevmode\llap{\theenmark.\enskip}}

\usepackage{natbib}
 \bibpunct[, ]{(}{)}{,}{a}{}{,}%
 \def\bibfont{\small}%
 \def\bibsep{\smallskipamount}%
 \def\bibhang{24pt}%

\newcommand{\Acal}{\mathcal{A}}
\newcommand{\Bcal}{\mathcal{B}}
\newcommand{\Dcal}{\mathcal{D}}
\newcommand{\Xcal}{\mathcal{X}}

\newcommand{\Tcal}{\mathcal{T}}
\newcommand{\Gcal}{\mathcal{G}}
\newcommand{\steptwo}{\eta}
\newcommand{\step}{\alpha}
\newcommand{\mubar}{\bar{\mu}}
\newcommand{\betabar}{\bar{\beta}}
\newcommand{\sigmabar}{\bar{\sigma}}
\newcommand{\Var}{\mbox{Var}}
\newcommand{\E}{\mathds{E}}
\newcommand{\Rhat}{\hat{R}}
\newcommand{\Dhat}{\hat{D}}
\newcommand{\nuhat}{\hat{\nu}}
\newcommand{\vhat}{\hat{v}}
\newcommand{\Vbar}{\bar{V}}
\newcommand{\vbar}{\bar{v}}
\newcommand{\bns}{\begin{eqnarray*}}
\newcommand{\ens}{\end{eqnarray*}}
\newcommand{\bn}{\begin{eqnarray}}
\newcommand{\en}{\end{eqnarray}}
\newcommand{\textwrap}{\mbox}
\newcommand{\textwrapbig}{\mbox}

\TheoremsNumberedThrough     
\ECRepeatTheorems

\EquationsNumberedThrough    

\begin{document}

\RUNAUTHOR{Al-Kanj and Powell}
\RUNTITLE{Approximate Dynamic Programming for Planning Autonomous Fleets of Electric Vehicles}
\TITLE{Approximate Dynamic Programming for Planning a Ride-Sharing System using Autonomous Fleets of Electric Vehicles}
\ARTICLEAUTHORS{
\AUTHOR{Lina Al-Kanj, Juliana Nascimento and Warren B. Powell}
\AFF{Operations Research and Financial Engineering Department, Princeton University, NJ, USA \EMAIL{\{lalkanj,jnascime,powell\}@princeton.edu}}
}

\ABSTRACT{
Within a decade, almost every major auto company, along with fleet operators such as Uber, have announced plans to put autonomous vehicles on the road. At the same time, electric vehicles are quickly emerging as a next-generation technology that is cost effective, in
addition to offering the benefits of reducing the carbon footprint. The combination of a centrally managed fleet of driverless vehicles, along with the operating characteristics of electric vehicles, is creating a transformative new technology that offers significant cost savings with high service levels.
This problem involves a dispatch problem for assigning riders to cars, a surge pricing problem for deciding on the price per trip and a planning problem for deciding on the fleet size. We use approximate dynamic programming to develop high-quality operational dispatch strategies to determine which car is best for a particular trip, when a car should be recharged, and when it should be re-positioned to a different zone which offers a higher density of trips. We prove that the value functions are monotone in the battery and time dimensions and use hierarchical aggregation to get better estimates of the value functions with a small number of observations. Then, surge pricing is discussed using an adaptive learning approach to decide on the price for each trip. Finally, we discuss the fleet size problem which depends on the previous two problems.
}

\SUBJECTCLASS{Transportation: Vehicle Routing, Dynamic Programming: Applications}
\maketitle

\section{Introduction}
Autonomous fleets of electric vehicles have emerged as a powerful combination of technologies and processes that have the potential of producing the biggest transformation of modern society since the introduction of the automobile (\citet{Galluci2016}).  This change is arising as a result of the emergence of three fundamental technologies: 1) the ability of mobile apps, pioneered by Uber, that allows riders and passengers to virtually self-manage a large fleet, 2) the maturation of autonomous control that will soon make driverless vehicles possible (estimates of the timeline for this range from  2025-2040), and 3) the emergence of electric vehicles (or cars), which represents the kind of technological change seen in air transportation with the emergence of gas turbine engines (\citet{Mui2016}). This technology, combining both electric cars and a central fleet operator, has the potential of transforming personal transportation, reducing operating and capital costs, reducing injuries and sharply reducing the need for parking in congested areas.  Fleet operators can also encourage ride sharing through price incentives, which reduces total trips.

The largest cost of an electric vehicle (EV) is the battery, but the goal of $300$+ mile ranges is due purely to the requirement that it cover the longest trip that an owner might require.  In a fleet, batteries can be much smaller since the fleet operator can draw on a much wider range of choices to find a good match for a customer. The challenge this introduces is that the fleet operator now has to manage the recharging process, which means thinking about which car is best for a particular rider given the origin, destination, and the battery level of the car.  By contrast, fleet operators today myopically assign drivers to riders without considering the downstream impact.

Carsharing systems have attracted attention in the literature (e.g., see \citet{Agatz2012,Li2014,Boyaci2015,Banerjee2018}). There are several variations in the modelled systems such as cars with drivers or driverless, gas or electric vehicles which introduces different challenges to the addressed problems. Autonomous cars can be active the whole day except for the recharging/refueling time. On the other hand, the recharging process of electric vehicles takes much longer time compared to gasoline cars (e.g. charging a 65kWh battery can take 45 minutes using an electric charger with a recharge rate of 300 miles/hour which is a relatively a fast recharger). Studies looking at shared electric autonomous vehicles are very limited~(\citet{Chen2016b}). All of this work has looked at a limited set of decisions that are required to run a fleet of vehicles and majority have worked with steady state systems (see Section~\ref{sec:litrev_risesharing}).

In this work, we provide a detailed model for a fleet of driverless cars taking into account many of the decisions required to operate a fleet in reality. These decisions are car-rider allocation, car relocation, parking or recharging taking into account the future trip demands. Our model optimizes across the fleet, both at a point in time and over time.  Optimizing over time means that we have to consider the impact of decisions now on an uncertain future where the model understands the time-of-day travel patterns, allowing it to produce behaviors where cars recharge primarily during off-peak periods and relocate to locations where trips are anticipated. The proposed framework can work for any stochastic trip distribution and is developed to handle any randomness at each point in time. Not only this, we are proposing these operations using a sequential optimization framework and try to approximate the optimal policy.

This paper makes the following contributions:
\begin{itemize}
\item[1.] The dispatch problem, where we have to optimize the assignment of cars to riders to optimize the use of available storage. It involves the management of the vehicles to meet the spatial and temporal characteristics of the trips, while simultaneously managing the process of recharging batteries, repositioning  and parking the vehicles. We solve this problem using approximate dynamic programming and adopt two strategies that prove critical to the success of the method.  We prove that the value functions are monotone in both time and the  battery charge level; using the monotonicity property accelerates convergence. We also use hierarchical aggregation to get better estimates of the value functions with a small number of observations.
\item[2.] The surge pricing problem, which offers the potential for smoothing out daily peaks.  We use adaptive learning techniques to guide this process in a real-time setting.
\item[3.] The fleet size problem, where we use the results of the dispatch and surge pricing problems to guide the design of the fleet size. We show that as the fleet size increases, the required battery size of the cars decreases.
\item[4.] We have developed a comprehensive simulator that simulates the area of New Jersey and also uses real data sets for trips across New Jersey. The simulator is scalable and can simulate any fleet size, area and trip distribution over any period of time. 
\end{itemize}


This paper is organized as follows. Section~\ref{sec:Lit_Review} presents a literature review on autonomous fleets of electric vehicles. Section~\ref{sec:Prob_Description} describes the problem and the decisions that are addressed in this work. Section~\ref{sec:Prob_Formulation} presents the mathematical model.  Section~\ref{sec:algorithmicstrategies} describes different policies for making the vehicle assignment decisions, including both a hybrid myopic policy with rules, as well as approximate dynamic programming. Section~\ref{sec:Surge_Pricing} presents the surge pricing problem and Section~\ref{sec:FleetSize_Problem} discusses the fleet size problem. Performance results of the various policies using real trip data sets are presented in Section~\ref{sec:results}. Finally, conclusions are drawn in Section~\ref{sec:conclusion}.

\section{Literature Review of Autonomous, Electric Vehicles}\label{sec:Lit_Review}
We begin by providing some background on the growth of autonomous transportation, and the emergence of electric vehicles.
\subsection{Autonomous Vehicles}
An autonomous vehicle is a car that is capable of sensing its environment and navigating without human input. These cars sense their surroundings using different techniques such as radar, GPS, odometry, and computer vision (\citet{Sivaraman2013}). Advanced control systems interpret the sensed information to identify appropriate navigation paths, as well as obstacles and relevant signage (\citet{Dokic2015}). The major anticipated advantages of driverless cars are:
\begin{itemize}
\item[1. ]Less traffic, lower accident rates and lower insurance costs: The potential reduction in traffic and deadly crashes caused by human-driver such as delayed reaction time or aggressive driving; thereby reducing the travel time and increasing the safety on the roads. Additional advantages could include higher speed limits, increased roadway capacity which result in an improved ability to manage traffic flow, combined with less need for traffic police (\citet{Luettel2012}). Also the vehicles' increased awareness could reduce car theft. All these factors contribute to decreasing the vehicle insurance (\citet{Fagnant2015}).
\item[2. ]Increased commuter flexibility and time efficiency: Relieving the commuters from driving where they can use the commuting time more efficiently for work or relaxation. This also removes the constraints on the commuter’s capability and age to drive where the commuter can be under age, over age, disabled or drunk~(\citet{Fagnant2015}).
\end{itemize}

Given all these benefits of autonomous vehicles, research on this technology dates back to the 1920s and 30s; however, the first autonomous car appeared in the 1980s, with Carnegie Mellon University's Navlab and ALV projects in 1984 (\citet{Kanade1986}). Since then, a number of major car manufacturers, ride-sharing companies and technology companies have developed working prototypes of driverless vehicles including Mercedes-Benz, General Motors,  Tesla Motors, Waymo and Uber. 

In 2015, Uber hired $40$ top scientists and researchers from Carnegie Mellon University to develop Uber's own autonomous capabilities (\citet{Mui2016}). In May 2014, Waymo revealed a new prototype of its driverless car, which had no steering wheel, gas pedal, or brake pedal, and was fully autonomous. As of March 2016, Waymo had test driven their fleet of cars in autonomous mode a total of 1.5 million miles with only 12 accidents (\citet{Associated-Press2015}) which are generally due to other drivers, a major challenge for driverless fleets. All this ongoing research indicates that autonomous cars are soon to become a reality.

\subsection{Electric Vehicles}
Electric vehicles have emerged over the past decade, and while total sales remain small, all the major manufacturers, along with a number of startups, have moved aggressively forward with this technology which, aside from the notable exception of the battery, is dramatically simpler than an internal combustion engine and transmission.  Electric vehicles have begun to catch up with conventional cars in terms of creature comforts, and have particularly benefited from improvements in battery technologies (\citet{Bruce2012}).  The major advantages of electric vehicles are:
\begin{itemize}
\item[1. ]Lower pollution: EVs have the potential of significantly reducing city pollution by having zero tail pipe emissions. With the U.S. energy mix using an electric car would result in a $30$\% reduction in carbon dioxide emissions (\citet{Faiz1996}).
\item[2. ]Lower production cost: Electric motors are mechanically very simple and often achieve $90$\% energy conversion efficiency over the full range of speeds. They can also be combined with regenerative braking systems to reduce the wear on brake systems which reduces the total energy requirement of a trip. They can be finely controlled and provide high torque from rest which removes the need for gearboxes and torque converters (\citet{Silberg2012}).
\item[3. ]Energy efficiency and resilience: EV energy efficiency is about a factor of $3$ higher than internal combustion engine vehicles. In addition, electricity can be produced from a variety of sources over the grid, allowing it to take advantage of changing commodity prices and evolving technologies (\citet{Guo2016}).
\item[4. ]Operating/Recharge cost: General Motors states that the Chevy Volt should cost less than $2$ cents per mile to drive on electricity, compared with $12$ cents a mile on gasoline at a price of \$$3.60$ a gallon (\citet{Valdes-Dapena2008}).
\item[5. ]Stabilization of the grid: Since EVs can be plugged into the electric grid when not in use, there is a potential for battery powered vehicles to cut the demand for electricity by feeding electricity into the grid from their batteries during peak use periods (\citet{Sortomme2012,Liu2013}). This vehicle-to-grid connection has the potential to reduce the need for new power plants.
\end{itemize}
Research on electric vehicles has started since the 19th century. In fact, in 1900, EVs formed $28$ percent of the cars on the road, but this technology was quickly overtaken by gasoline-based engines.  During the last few decades, significant advancement and experimentation has been done in the EV industry. As of September 2014, EVs have been available in some countries for retail customers produced by Mitsubishi, Nissan Leaf, Renault Fluence, Ford Focus Electric, Tesla Model S among others. As of early December 2015, Leaf sold $200,000$ EVs worldwide followed by the Tesla Model S with global deliveries of about $100,000$ EVs (\citet{Cobb2015}).

\subsection{Autonomous Fleets of Electric Vehicles}
The switch to EVs, the use of car sharing services, and the arrival of autonomous vehicles are all expected to grow substantially by 2030, potentially making it cleaner and easier to navigate cities around the world, Bloomberg New Energy Finance (BNEF) said in a new report (\citet{Galluci2016,McDonald2016}). BNEF and consulting firm McKinsey \& Company looked at how these three developments could play out in $50$ cities around the world. They found that each city's demographics and existing transportation policies would affect the technological adoption. In densely populated cities, the analysts found that EVs would replace a large share of gas-powered cars. However, in more sprawling metro areas, they predicted a rise in autonomous cars. Thus, the driverless EVs fleet technology is seen as one of the promising future technologies for the benefit of people moving around in cities, in terms of environmental impacts, ease of transport and lower operating, capital, insurance and parking costs (\citet{Chen2015}). \citet{Clewlow2017} present a detailed study on how ride-hailing services affects commuter behaviour.

While there is an obvious and compelling benefit of this new vision in the wholesale reduction in the burning of oil, the case for autonomous fleets of EVs is entirely economic.  These include in addition to the benefits mentioned above:
\begin{itemize}
  \item[1.]Reduced operating and capital cost, resulting from higher utilization (with an estimated increase from $5$ percent utilization to as much as $30$ percent), and the ability to use smaller cars.  When driverless cars are used as cabs, then the operating cost is significantly reduced by removing the cost of the driver which accounts for the highest cost for a ride fare; this factor also contributes to new business models that are cheaper than car ownership which is important for the population with low income. Also car sharing reduces the total number of cars to satisfy the transportation cost resulting in lower capital and operating costs (\citet{Litman2014,Spinoulas2015}).
   \item[2.] Reduced parking costs; suburban businesses often require parking lots that are $8-10$ times the footprint of an office building Roads often have to be built to accommodate parking. The area required for vehicle parking would also be cut down, as these cars would be able to go where space is more readily available which reduces parking cost (\citet{Anderson2016}). Parking represents the top reason that urban ride-hailing users substitute a ride-hailing service in place of driving themselves (37\%)~\citet{Clewlow2017}.
   \item[3.] Cost benefits have been estimated to range from \$$2,000$ to \$$5,000$ annually (\citet{Fagnant2015}),  but these estimates are not able to account for the transformations to urban and suburban forms with the dramatic reduction of space for parking.  These estimates also do not fully account for the benefits of a potentially dramatic reduction in deaths and injuries due to traffic accidents.  In (Burns 2013), it is estimated that driverless EVs could be more than $70$\% cheaper for a citizen of a city like Ann Arbor and would require residents to invest less than one-fifth of the amount needed to own their cars. Also, about $80$\% fewer shared EVs would be needed than personally owned vehicles to provide the same level of mobility, with less investment but increase travel per vehicle by up to $75$\% (\citet{Schoettle2015}).
\end{itemize}

\subsection{Ride Sharing Systems}\label{sec:litrev_risesharing}
We start by giving an overview about ride-sharing taxis with drivers and then autonomous will follow. In~\citet{Lin2012},  a routing optimization model for taxi ride-sharing is proposed and solved via a simulated annealing algorithm to address a static, multiple-cars problem with time windows. In~\citet{Sti2017}, each driver  has a specific itinerary and is willing to pick up and drop off riders en route. A mixed-integer linear programming model has been developed for the single driver multiple riders matching problem whereas a heuristic solution is used for the multiple drivers problem. A queueing network model that focuses on empty-car routing is presented in~\citet{Braverman2017} to control car flow in the network to optimize the availability of empty cars when a passenger arrives. \citet{Banerjee2018} propose a queueing network model for designing a dispatch policy, where the platform can choose which vehicle to assign when a customer request comes in, and assume that this is the exclusive control lever available. Their performance measure is minimizing the proportion of dropped demand in steady state.

There is also a literature tackling ride sharing in autonomous systems. \citet{Wang2006} proposed a dynamic fleet management algorithm for shared fully automated vehicles based on queuing theory for simulating a dispatch policy.   In \citet{Kornhauser2013} autonomous taxi stands are placed in every half mile by half mile zone across New Jersey, and passengers walk to taxi stands rather than allowing AVs to relocate. \citet{ITF2015} looked at the application of shared and self-driving vehicles in Lisbon, Portugal where they simulated a heuristic dispatch policy. This study also looked at the impact of electrifying shared self-driving vehicles, assuming an electric range of 175 kilometers (108 miles) and a recharge time of 30 minutes, and found that the fleet would need to be 2\% larger. \citet{Fagnant2014} presented an agent-based model for shared autonomous vehicles which simulated environmental benefits of such a fleet as compared to conventional vehicle ownership. Simulation results indicated that each autonomous car can replace 11 conventional private owned vehicles, but generates up to 10\% more travel distances.

Charging/refueling in a fleet of shared self-driving vehicles was missing in all of the prior studies mentioned  except \citet{Fagnant2014} and \citet{Chen2016b}. \citet{Fagnant2014} modeled the logistics of refueling by assuming the autonomous car could refuel at any location within the grid with a fixed service lag time. \citet{Chen2016b} builds on the work of~\citet{Fagnant2014} where a similar dispatch framework is used to assign the locations of charging stations in a heuristic manner;  a recharging station was generated if a car needs to recharge and there is no station within a certain range.


\section{Problem Description}
\label{sec:Prob_Description}
In this work, we assume that there is an autonomous fleet of $N^{\mbox{\footnotesize cars}}$ electric vehicles that serves a set of trips reflecting the spatial and temporal demand patterns for New Jersey over the course of a single day. The set of trip distribution is not uniform, typically there are peaks in the morning and afternoon hours where the number of trips is much higher compared to the other hours of the day. The aim of this work is to maximize the revenue of the fleet and study the economics of such a system. In order to achieve this, we will look at three decision problems that are detailed below.

\begin{itemize}
\item[\textbf{1.} ]\textbf{The dispatch problem:} optimizes the assignment of cars to riders to maximize revenue which is a high-dimensional stochastic dynamic program. Currently, transportation companies use relatively simple rules to assign an EV to a rider; for example, Uber checks up to eight closest drivers to see if one agrees to pick up a requested trip.

In this work, car assignment, battery recharging, repositioning and parking a car are based on an optimized policy that takes into account all the available travel requests, their destinations, the available battery charges of the cars and the value of being at a certain destination. For example, it is more efficient to assign a car with a low battery level to a destination that is already close to a charging station. Thus, knowing the charge level of each car and the trip required for each rider, what is the best vehicle to assign to a rider? Moreover, we need to decide not just when to recharge, but how long to keep a car at a recharging station.  Particularly challenging is planning battery recharging to maximize availability during peak periods. Also, car repositioning to where nearby trips are incoming maximizes trip coverage and minimizes the customer wait time; this depends on the time of the day.



In this model, we consider simultaneously the decision of which car to assign to each rider, which cars to recharge, and which to reposition empty to another zone.  We propose an approximate dynamic programming algorithm to capture the impact of decisions now on the future.

\item[\textbf{2.} ]{\textbf{The surge pricing problem:}} As Uber realized early in its evolution, it is necessary to use pricing to balance supply and demand, so that there is always a good balance of cars and riders.  The market response to price has to be learned dynamically, and depends on both location and~time which we model using an adaptive learning framework. 

\item[\textbf{3.} ]{\textbf{The fleet size problem:}}
The aim of any fleet operator is to decide on its fleet size so that it covers a certain percentage of the trips evolving over the whole day. The challenge here is understanding the marginal value of cars (taking peak periods into account); this is guided by the marginal value functions obtained using the dispatch and surge pricing problems.
\end{itemize}

\section{Problem Formulation}\label{sec:Prob_Formulation}
We model the problem using the language of dynamic resource management (see \cite{PoShSi01}) where cars are ``resources.'' The tasks of the cars include fulfilling trips, repositioning, recharging and staying in the same place. We present the mathematical model of the fleet management problem as a Markov decision process with discrete time and discrete state space. It consists of five core components:  state variables, decision variables, exogenous information, the transition function and the objective~function. 

\subsubsection*{\textbf{The State Variables} }
The state consists of physical state variables, information state variables, and belief state variables.  The physical state, $S^R_t$, describes the state of all the cars, as well as the state of customers wishing to be served. The information state includes any information that is known deterministically before a decision epoch and the belief state, $B_t$, contains distributional information about unknown parameters such as prices. In this work, the state of the system includes both a physical state and a belief state, i.e., $S_t=(S^R_t,B_t)$. We start first by modelling the system using the physical state only and then we discuss the belief state in Section~\ref{sec:Surge_Pricing} which models surge pricing. If the prices per zone are known at each point in time, then there is only a physical state in the system, i.e., $S_t=S^R_t$.

A car is described by an attribute vector $a$, which captures the time of availability/arrival, location of the car, and battery level.  This can be written as:
\bns
a &=& \left(\begin{array}{c} a_1 \\a_2 \end{array} \right)
= \left(\begin{array}{c}  \mbox{location}
\\ \mbox{battery Level}
\end{array} \right) = \left(\begin{array}{c} z
\\  l
\end{array} \right)\\
\vspace{0.02in} \nonumber \\
\mathcal{A} &=& \mbox{Set of all possible car attribute vectors $a$.}
\ens
The location $z$ corresponds to a two dimensional vector representing an $x$ and $y$ coordinate. We represent physical location by dividing a region (such as the state of New Jersey) into small square zones with width $w^{\mbox{zone}}$, where we assumed that $w^{\mbox{zone}}=0.5$ miles. Let $\mathcal{Z}$ be the set of all zones of the given area and $b^{\mbox{\footnotesize size}}_i$ be the maximum battery size of car $i$ which is discretized into $l^d$ levels.

Similarly, let $b$ be the vector of attributes of a trip, including elements such as time, origin and destination. Let $\Bcal$ be the space of all trips. The vector of attributes of a trip is given by
\bns
b &=& \left(\begin{array}{c} b_1 \\ b_2 \end{array} \right)
= \left(\begin{array}{c}\mbox{origin}  \\ \mbox{destination}
\end{array} \right) = \left(\begin{array}{c}  z_o \\ z_d
\end{array} \right)\\
\vspace{0.10in} \\
\Bcal &=& \textwrap{Set of trip attribute vectors $b$.}
\ens
The time of trip requests is continuous, but the decision epoch is discrete as will be explained later. If the decision epoch is at time $t$, then all trips arriving between $t-1$ and $t$ are collected and assigned at time $t$. The origin and destination of the trips, $z_o$ and $z_d$ are also in the set $\mathcal{Z}$.

Let $a_t$ be the attribute vector of a car at time $t$.  We model the state of all the cars using the resource state vector, which is defined using
\bns
R_{ta} &=&  \textwrap{The number of resources with attribute vector $a$ at time $t$.}\\
R_t    &=&  \textwrap{The resource state vector at time $t$.}\\
       &=&  \left(R_{ta}\right)_{a \in \Acal}.
\ens
We then let $D_{tb}$ be the number of trips with attribute $b$, and let $D_t = (D_{tb})_{b\in\Bcal}$.
Note that, $R_t$ contains all cars; however, we only consider the cars that are available to be allocated for a decision. The cars that are busy fulfilling trips or recharging are not considered to be reassigned for a decision.
The physical system state vector is then given by
\bns
S^R_t = (R_t,D_t).
\ens
If the prices are fixed for all zones then there is no belief state and the system only has a physical state; in this case, the state $S_t$ is equivalent to the physical state $S^R_t$. We measure the state $S_t$ just before we make a decision.  These {\it decision epochs} are modeled in discrete time $t = 0, 1, 2, \ldots, T$, but the physical process occurs in continuous time.


\subsubsection*{\textbf{The Decision Variables} }
Each car with attribute $a_t$ can be acted on by one of the following decisions: 1) assign it to one of the trips in $D_t$ within its pickup range which is set by a parameter $D^R$, 2) move it empty to one of its neighboring zones, 3) recharge it in its current zone or 4) hold it in its current zone. In this work, we assume there is a recharging station in every zone whereas optimal recharging station placement is left as a future work. What differs in this case is that a car should travel a few miles to reach a recharging station which consumes some of its battery.  Decisions are described using
\bns
d       &=& \textwrapbig{An elementary decision,}\\
\Dcal^L   &=& \textwrapbig{The set of all decisions to cover a type of trip, where an element $d\in\Dcal^L$ represents a} \\
&& \hspace{0cm} \textwrapbig{decision to cover a trip of type $b_d\in\Bcal$,}\\
d^{\mbox{\footnotesize empty}} &=& \textwrapbig{The decision to move empty to one of neighboring zones,}\\
\Dcal^{\mbox{\footnotesize empty}}   &=& \textwrapbig{The set of all decisions to move empty where $d^{\mbox{\footnotesize empty}}\in\Dcal^{\mbox{\footnotesize empty}}$,}\\
d^{\mbox{\footnotesize recharge}} &=& \textwrapbig{The decision to recharge at one of the recharging stations,}\\
\Dcal^{\mbox{\footnotesize recharge}}   &=& \textwrapbig{The set of all decisions to recharge where $d^{\mbox{\footnotesize recharge}}\in\Dcal^{\mbox{\footnotesize recharge}}$,}\\
d^{\mbox{\small stay}} &=& \textwrapbig{The decision to stay,}\\
\Dcal &=& \Dcal^L \cup \Dcal^{\mbox{\footnotesize empty}} \cup \Dcal^{\mbox{\footnotesize recharge}} \cup d^{\mbox{\small stay}},\\
x_{tad} &=& \textwrapbig{The number of times decision $d$ is applied to a resource (car) with attribute vector $a$} \\
&&\textwrapbig{at time $t$,}\\
x_t     &=& (x_{tad})_{a \in \Acal, d \in \Dcal}.
\ens
The decision variables $x_{tad}$ have to satisfy the following constraints:
\bn
\sum_{d\in \Dcal} x_{tad} &=& R_{ta}  ~~~~\forall a \in \Acal,      \label{eq:constr1}\\
\sum_{a \in \Acal} x_{tad} &\le& D_{tb_d} ~~~~ \forall d \in \Dcal^L,      \label{eq:constr2}\\
x_{tad} &\ge& 0 ~~~~ a \in \Acal, d \in \Dcal.     \label{eq:constr3}
\en

Equation \eqref{eq:constr1} captures flow conservation for cars (we cannot assign more than we have of a particular type). Equation \eqref{eq:constr2} is flow conservation on trips (we cannot assign more cars to trips of type $b_d$ than there are trips of this type).  We let $\Xcal_t$ be the set of all $x_t$ that satisfy equations \eqref{eq:constr1} - \eqref{eq:constr3}.  The feasible region $\Xcal_t$ depends on $S_t$.  Rather than write $\Xcal(S_t)$, we let the subscript $t$ in $\Xcal_t$ indicate the dependence on the information available at time $t$.  We assume that decisions are determined by a decision function denoted
\bns
X_t^\pi(S_t) &=& \textwrap{A function that determines $x_t \in \Xcal_t$ given $S_t$, where $\pi\in\Pi$,}\\
\Pi        &=& \textwrap{A set of decision functions (or policies).}
\ens

\subsubsection*{\textbf{The Exogenous Information }} This represents the new information arriving between time $t-1$ and $t$. In this model, there are two types of exogenous information processes: updates to the attributes of a car, and new trip demands. Define

\bns
\Rhat_{ta} &=& \textwrapbig{The change in the number of cars with attribute $a$ due to information arriving between}\\
             && \textwrapbig{time $t-1$ and $t$. This might capture cars entering and leaving service, as well as random}\\
              && \textwrapbig{delays in the arrival of cars moving from one location to another.}\\
\Dhat_{tb} &=& \textwrapbig{The number of new trips that first became known to the system with attribute $b$ between}\\
             && \textwrapbig{time $t-1$ and $t$.}
\ens
For example, $\Dhat_{tb} = +1$ if we have a new trip order with attribute vector $b$.  If a car attribute randomly changed from $a$ to $a'$ (arising, for example, from  travel delays, cars entering/leaving the system, equipment failure, etc.), we would have $\Rhat_{ta} = -1$ and $\Rhat_{ta'} = +1$.  We let
$$W_t = (\Rhat_t,\Dhat_t)$$
be our generic variable for new information.  We view information as arriving continuously in time, where the interval between time instant $t-1$ and $t$ is labeled as time interval $t$.  Thus, $W_1$ is the information that arrives between now ($t=0$) and the first decision epoch ($t=1$).

\subsubsection*{\textbf{The Transition Function} } It represents how the system evolves over time. 
If decision $d$ is applied to a car with attribute $a$, this changes the car attribute vector to
\bns
a' = a^M(a,d).
\ens
We model the transition function deterministically, which means that $a'$ is the attribute vector that we think results from a decision, but before any new information has arrived.  So, if we decide to move a car from location $z_1$ with battery level $l_1$ to location $z_2$ consuming a battery of $\triangle l$, then immediately after the assignment, this would be a car with the attribute that we expect it to be at location $z_2$ with battery level $l_1-\triangle l$. For algebraic purposes, define
\bns
\delta_{a'}(a,d) &=& \left\{
\begin{array}{ll}
1, & \mbox{if $a^M(a,d) = a'$,} \\
0, & \mbox{otherwise.}
\end{array}
\right.
\ens

When the car repositions, its new set of attributes are the time to reach the new location, the new location it moved to and the battery level remaining after reaching the new location. When the car recharges, the new attributes are the time it took for recharging which is a decision variable and the new battery level after recharging. When the car is held in the same location, the only attribute that changes is the time of the availability of the car which is typically the next decision~epoch.

We now define the {\it post-decision} resource vector, which is the resource vector after we make a decision, but before any new information arrives.  This can be written
\bn
R^x_{ta'} &=& \sum_{a \in \Acal} \sum_{d \in \Dcal} \delta_{a'}(a,d) x_{tad}.       \label{eq:transition1}
\en

Finally, the next pre-decision resource vector contains all the set of cars that become available at time $t+1$ which is given by
\bn
R_{t+1,a} = R^x_{ta} + \Rhat_{t+1,a}.    \label{eq:transition2}
\en
It is more conventional in stochastic dynamic systems to write the transition from $R_t$ to $R_{t+1}$. Capturing the post-decision resource vector provides significant computational advantages, as we illustrate~later.

We make the assumption that all demands are either satisfied or lost as travelers find other means of transportation.  We could introduce backlogging, in which case we would add an attribute to the vector $b$ that captures how long a trip has been waiting.  For now, we are going to assume that all trips are new trips, which means that $D_{t+1} = \Dhat_{t+1}$.

We denote the transition function~by
\bn
S_{t+1} = S^M(S_t,x_t,W_{t+1})
\en
where $S^M()$ governs the transition from pre-decision state $S_t$ to pre-decision state $S_{t+1}$. This transition function can be divided into two consecutive functions $S^x_t = S^{M,x}(S_t,x_t)$ and $S_{t+1} = S^{M,W}(S^x_{t},W_{t+1})$ where $S^{M,x}$ is the transition function from a pre-decision state to a post-decision state and $S^{M,W}()$ is the transition function from a post-decision state to a pre-decision state.

\subsubsection*{ \textbf{The Objective Function} }
When a decision $d$ is applied to a car with attribute $a$ at time $t$, it produces a contribution $c_{tad}$. Taking a trip generates revenue, whereas moving empty or staying in the same place generates zero revenue. As in reality, we assume that a car is paying money at the time of recharging which depends on the amount of energy placed in the battery that is consumed as the car is performing the various tasks. This contribution is given~by:
\bn
c_{tad} &=& \left\{
\begin{array}{ll}
\rho^m + p^m_b\cdot \delta(b), & \mbox{if decision $d$ is to take a trip with attribute $b$} \\
-(\rho^r + p^{r}\cdot \eta^r\cdot \triangle t^{\mbox{\footnotesize{recharge}}}), & \mbox{if decision $d$ is to recharge}\\
0, & \mbox{if decision $d$ is to stay or move empty}
\end{array}\label{eq:rewardfunction}
\right.
\en
where $\rho^m$ is the base fare for taking a trip, $p^{m}_b$ is the revenue (price) per mile which depends on the attributes of the trip $b$ such as the zone where the trip originates and the time of the day, $\delta(b)$ is the distance of the trip. $\rho^r$ is the base fare for recharging, $p^{r}$ is the energy cost per mile (which is equivalent to the cost paid to recharge the battery with some kWh that cover a mile), $\eta^r$ is the rate of recharging in miles/seconds and $\triangle t^{\mbox{\small{recharge}}}$ is the total time in the recharging process.  Based on the decision, the attribute of the car changes from $a$ to $a'=a^M(a,d)$. If we are not applying surge pricing, then the price, $p^{m}_b$, can be considered constant for all trips.

If we assume that the contributions are linear, the contribution function for period $t$ would be given by
\bn
C_t(S_t,x_t)  &=& \sum_{a \in \mathcal{A}} \sum_{d \in \Dcal} c_{tad}x_{tad}.            \label{eq:fmplinearcosts}
\en

The optimal policy maximizes the expected sum of contributions,  over all the time periods
\bn
F^*_0(S_0) = \max_{\pi \in \Pi} \E \left\{\sum_{t = 0}^{T} C\left(S_t,X_t^\pi(S_t)\right)| S_0\right\}.         \label{eq:objectivefunction}
\en
In the next section, we describe the policies we tested for solving \eqref{eq:objectivefunction}.

\section{Designing Policies}\label{sec:algorithmicstrategies}
In this section, we discuss the policies used to solve the formulated problem. There are two strategies for designing policies:
\begin{itemize}
\item[\textbf{1.}] \textbf{Policy search} - This is where we search within a (typically parametric) class of policies.  This can be done in two ways:
\begin{itemize}
\item[\textbf{a)}] \textbf{Policy function approximations (PFAs)} - These are analytic functions that map states to actions, which may be a linear model, a rule-based function, or a neural network.
\item[\textbf{b)}] \textbf{ Parametric cost function approximations (CFAs)} - This is when we minimize a parametrically modified cost function, subject to (possibly) parametrically modified constraints.
\end{itemize}
\item[\textbf{2.}] \textbf{Lookahead policies} - This is when we model the downstream impact of a decision made now on the future.  An optimal policy can be written as
\begin{eqnarray}
X^*_t(S_t)=\arg\max_{x_t\in{\mathcal{X}_t(S_t)}}\left(C(S_t,x_t)+\mathbb{E}\left\{\max_\pi \mathbb{E}^\pi\left\{ \sum_{t'=t+1}^T C(S_{t'},X^\pi_{t'}(S_{t'}))|S_{t+1}\right\}|S_t,x_t\right\} \right), \label{eq:opt_policy}
\end{eqnarray}
While the expectations and the maximization over policies is typically intractable, this equation serves as the basis for two classes of approximations:
\begin{itemize}
\item[\textbf{a)}] \textbf{Direct lookahead policies} - Here, we try to directly solve \eqref{eq:opt_policy} by replacing the complex expectations by using a simpler model (such as deterministic), and by replacing the search over policies with something more tractable. This is how popular navigation systems plan paths. One can also use stochastic lookahead models using Monte Carlo tree search for example (see \citet{Jiang2017} and \citet{Al-Kanj2016}).
\item[\textbf{b)}] \textbf{Value function approximations (VFAs)} -  Widely known as approximate dynamic programming (ADP)
or reinforcement learning, value function approximations replace the lookahead model
with a statistical model of the future that depends on the downstream state resulting from
starting in state $S_t$ and making decision $x_t$.
\end{itemize}
\end{itemize}
For this complex problem, we will be using a mixture of policies to handle decisions such as whether or not to charge the vehicle, assignment of the vehicle to a customer, and the price to charge for a trip. First, we approach the problem using a myopic policy based on a parameterized CFA policy as presented in Section~\ref{sec:myopic_policy}. Then, a lookahead policy based on VFAs is presented in Section~\ref{sec:lookahead_policy} which takes into account the downstream value of the decision. In this work, a lookahead policy based on value function approximations is used to solve the problem as discussed in Section~\ref{sec:introadp}. Finally, Section \ref{sec:hieragg} shows an aggregation algorithm for improving the value functions even with a small number of observations for a zone (to measure the value function of a zone at a given point in time, we need a car there).

\subsection{A Parameterized Myopic Policy}\label{sec:myopic_policy}
One policy for solving this problem is the myopic policy, i.e., by taking a decision without measuring its impact on the future. This policy  combines a linear program for assigning cars to riders, with a parameterized rule that determines when a car should be recharged (e.g. recharge if charge level is less than $\theta^{\mbox{\footnotesize thr}}$ which needs tuning). Hence, the proposed myopic policy is equivalent to using a parameterized CFA policy (solving a parameterized optimization problem). The problem is formulated as:
\bn
&&\max_{x_t\in\Xcal_t} \sum_{a \in \Acal} \sum_{d \in \Dcal} c_{tad}x_{tad}\nonumber\\
&&\mbox{subject to}\nonumber\\
&&x_{tad^{\mbox{\footnotesize{recharge}}}} = R_{ta}, \mbox{if~} a_3 < \theta^{\mbox{\footnotesize{thr}}}*b^{\mbox{\footnotesize size}} \label{eq: constr4}
\en
where $a_3$ represents the third attribute of the car which corresponds to its battery level. Constraint (\ref{eq: constr4}) forces the car to recharge if its battery level drops below a certain threshold $\theta^{\mbox{\footnotesize{thr}}}$. The feasible set $\Xcal_t$ is defined by constraints (\ref{eq:constr1}), (\ref{eq:constr2}) and (\ref{eq:constr3}).  Given a solution $x_t = X^\pi_t(S_t)$, we then use the transition functions to find $(R_{t+1},D_{t+1})$ by sampling $\Rhat_{t+1}(\omega)$ and $\Dhat_{t+1}(\omega)$.

Since this policy does not consider future events or patterns, it is unable to recognize that the system might be entering a peak period.  As a result, a car with a battery that is 40 percent charged during an offpeak period (when it can easily recharge) might not hit the recharge threshold until it is in the middle of an afternoon peak.

\subsection{Policies Based on Value Function Approximations}\label{sec:lookahead_policy}
A lookahead policy is particularly useful for this problem since it captures the impact of the decision taken now on the future. In principle, we can solve equation (9) using Bellman's optimality principle which allows us to write
\bn
X^*_t(S_t)=\arg\max_{x_t \in \Xcal_t}\left(C_t(S_t,x_t) + \mathbb{E}\left\{V_{t+1}(S_{t+1})|S_t,x_t\right\}\right) \label{eq:opt_policy2}
\en
where $S_{t+1} = S^M(S_t,x_t,W_{t+1})$. Solving equation \eqref{eq:opt_policy2} encounters three curses of dimensionality: the state vector $S_t$ (which has dimensionality: $H =\binom{|\mathcal{Z}|\cdot l^d + K - 1}{K}\cdot\binom{|\mathcal{Z}| + |D_t| - 1}{|D_t|}$, which can be extremely large), the outcome space (the expectation is over a vector of random variables measuring $\binom{|\mathcal{Z}|\cdot l^d  + K - 1}{K}$), and the action space (the vector $x_t$ is dimensioned $H\times |D_t|$). Table~\ref{tab:complexity}  illustrates how quickly the problem size grows if we were to use the traditional methods of discrete dynamic programming. The problem size addressed in this work is much larger than the examples presented in Table~\ref{tab:complexity}; it has an area formed of $22000$ locations, a battery level discretized into $20$ levels, and  a fleet size of $1500$ cars; hence, the solution is intractable for this network configuration.

\begin{table} [tb]
\begin{center}
\begin{tabular}{|c|c|c|c|c|} \hline
$|\mathcal{Z}|$   &  $l^d$    &  $N^{\mbox{\footnotesize cars}}$  & H \\ \hline \hline
200      &    10    &   1     & 2000  \\ \hline
200     &    10    &   5   & $2.6800\times10^{14}$\\ \hline
2000    &    10    &   100   & $1.7390\times 10^{272}$ \\ \hline
2000    &    20    &  100  & $1.9485\times 10^{302}$ \\ \hline
\end{tabular}
\end{center}
\vspace{0.2cm}
\caption{Outcome State space size versus the various parameters including area size, battery levels, number of cars at time $t$.}
\label{tab:complexity}
\end{table}
To solve equation \eqref{eq:opt_policy2}, we can often remove the expectation by replacing the value function of $S_{t+1}$ by the value function around the  post-decision state $S^x_t$,   which is a deterministic function of $S_t$ and $x_t$, giving
\bn
X^*_t(S_t)=\arg\max_{x_t \in \Xcal_t}\left(C_t(S_t,x_t) + V^x_{t}(S^x_t)\right) \label{eq:opt_policy3}
\en
We still need to design an approximation of the value of being in the post-decision state that scales to high dimensional decision vectors. 

Section \ref{sec:introadp} provides a sketch of a basic approximate dynamic programming algorithm for approximating the solution of \eqref{eq:opt_policy3}. Section \ref{sec:valueestimation} describes how we update the value function.

\subsection{An Approximate Dynamic Programming Algorithm}
\label{sec:introadp}
Computing the value function of the post-decision state $V^x_t(S^x_t)$ exactly can be very hard despite the dimensionality reduction induced. For this reason, we use the framework of approximate dynamic programming to replace the value function with an approximation (\citet{Wickens2000,Powell2011,Bertsekas2011}). Our approach requires merging math programming with the techniques of machine learning used within approximate dynamic programming.  

We solve \eqref{eq:opt_policy2} by breaking the dynamic programming recursions into two steps:
\bn
V_t(S_t) &=& \max_{x_t \in \Xcal_t} \big(C(S_t,x_t) + V^x_t(S^x_t) \big),   \label{eq:bellman2}\\
V^x_{t}(S^x_{t}) &=& \E \left\{ V_{t+1}(S_{t+1}) \mid S^x_{t} \right\}, \label{eq:bellman1}
\en
where $S_{t+1} = S^{M,W}(S^x_{t},W_{t+1})$ and $S^x_t = S^{M,x}(S_t,x_t)$.
%

The basic algorithmic strategy works as follows.  At iteration $n$,  a sample path $\omega^n$ is generated which determines the exogenous information $W_{t+1}(\omega^n)$ that we observe after making decision $x^n_t$. Then, the next pre-decision state $S^n_{t+1}$ is computed using
\bns
S^n_{t+1} = S^{M,W}(S^{x,n}_{t},W_{t+1}(\omega^n)).
\ens
There are several ways to perform updates of value functions. The most popular is based on approximate value iteration, where we compute an estimate $\vhat^n_t$ of the value of being in a particular state $S^n_t$ at time $t$ during the $n^{\mbox{th}}$ iteration. The estimate $\vhat^n_t$ can be computed during a simple forward pass using
\bn
\vhat^n_t &=& \max_{x_t \in \Xcal^n_t(\omega^n)}\big(C_t(S^n_t,x_t) + \Vbar^{n-1}_t(S^{M,x}(S^n_t,x_t))\big), \label{eq:vhato}
\en
and assume that $x^n_t$ is the value of $x_t$ that solves \eqref{eq:vhato} which is used to compute the post-decision state $S^{x,n}_t = S^{M,x}(S^n_t,x^n_t)$ to continue the process.



We use a value function approximation that is linear in the resource vector $R_t$ which can be written as
\bn
\Vbar^{n}_t(S^x_t) &=& \Vbar^{n}_t(R^x_t)  \label{eq:fmpvfaapprox}\\
&=& \sum_{a' \in \Acal} \vbar^n_{t+\tau(t,a,d),a'} \sum_{a \in \Acal} \sum_{d \in \Dcal} \delta_{a'}(a,d) x_{tad},    \label{eq:fmpvfaapprox2}
\en
where $\tau(t,a,d)$ is the travel time for a car with attribute $a$ when applying decision $d$ at time $t$. When applying decision $d$ to a car with attribute $a$, its new attribute is $a'=a^M(a,d)$ and its new time of availability (time of arrival) is $t+\tau(t,a,d)$.  Equation (\ref{eq:fmpvfaapprox2}) is obtained by using the state transition equation (\ref{eq:transition1}) and $\vbar^n_{t+\tau(t,a,d),a'}$ is the marginal value of a car with attributes $a'$ at time $t+\tau(t,a,d)$ at iteration $n$. More specifically, $\vbar^n_{ta}$ means that if we drop a car in a zone with coordinates $a_1$ and battery level $a_2$, then the marginal contribution of that car from time $t$ up to the end of the simulation horizon $T$ is approximated by $\vbar^n_{ta}$.

This enables us to write the policy as
\bn
X^{\pi}_t(S_t) &=& \arg\max_{x_t \in \Xcal_t} \left(\sum_{a \in \Acal} \sum_{d \in \Dcal} c_{tad}x_{tad} + \sum_{a' \in \Acal} \vbar_{t+\tau(t,a,d),a'} \sum_{a \in \Acal} \sum_{d \in \Dcal} \delta_{a'}(a,d) x_{tad}\right)   \nonumber\\
&=& \arg \max_{x_t \in \Xcal_t(\omega)} \sum_{a \in \Acal} \sum_{d \in \Dcal} \left( c_{tad} +  \sum_{a' \in \Acal} \vbar_{t+\tau(t,a,d),a'} \delta_{a'}(a,d)\right) x_{tad}.  \label{eq:linvfadecfunc2}
\en
Recognizing that $\sum_{a'\in\Acal}\delta_{a'}(a,d) = \delta_{a^M(a_t,d_t)}(a,d) = 1$, we can write \eqref{eq:linvfadecfunc2} as
\bn
X^{\pi}_t(S_t) = \arg \max_{x_t \in \Xcal_t(\omega)} \sum_{a \in \Acal} \sum_{d \in \Dcal} \big( c_{tad} +  \vbar^{n-1}_{t+\tau(t,a,d),a^M(a,d)}\big) x_{tad}.  \label{eq:linvfadecfunc3}
\en
Equation \eqref{eq:linvfadecfunc3} says that the best decision for a car  depends on the immediate reward at time $t$ and on the value of the car after the decision is taken, i.e., the value of its new attributes. Note that we are not directly  imposing a constraint for recharging as in the myopic case. The recharging decision is an outcome of the optimization problem; that is, if the value of recharging determined by $\left(c_{tad} +  \vbar^{n-1}_{t+\tau(t,a,d),a^M(a,d)}\right)$ is better than any other decision, the car will recharge. The main motivation for recharging is the value function of the car with a higher battery level. 

The complexity for solving \eqref{eq:linvfadecfunc3} is equivalent to solving the original myopic problem if the value functions are known. The challenge is in getting these value functions which have to be solved iteratively in order to estimate them. This will be discussed in the following section.


\subsection{Value Function Updates}
\label{sec:valueestimation}
We use a classical ADP algorithm known as approximate value iteration, but adapted for the setting where we are estimating the marginal value of being in a state (see \citet{Powell2011}).  The method using a pure forward algorithm, where updates are performed as we step forward in time.   At iteration $n$, a sample path $\omega^n$ is generated which determines $\Rhat^n_t = \Rhat_t(\omega^n)$ and $\Dhat^n_t = \Dhat_t(\omega^n)$.

Assume that the previous decision (at time $t-1$) gave the post-decision state $S^{x,n}_{t-1}$.  Following the sample path $\omega^n$ generates state $S^n_t = S^{M,W}(S^{x,n}_{t-1},W_t(\omega^n))$ which determines the feasible region $\Xcal^n_t$. The objective function in equation \eqref{eq:linvfadecfunc3} is computed using value functions from iteration $n-1$. The decision at time $t$ is obtained by solving
\bn
F_t(S^n_t) &=& \max_{x_t \in \Xcal^n_t}\big(C_t(S^n_t,x_t) + \Vbar^{n-1}_t(S^x_t)\big), \label{eq:choosex}
\en
where $S^x_t = S^{M,x}(S^n_t,x_t)$.  Let $x^n_t$ be the value of $x_t$ that solves \eqref{eq:choosex}.  Note that $R^{n}_{t}$ affects \eqref{eq:choosex} through the flow conservation constraint \eqref{eq:constr1}. The dual variable, $\vhat^n_{t,a}$, for the flow conservation constraint \eqref{eq:constr1} represents the marginal value of an additional car and is given by:
\bn
\vhat^n_{t,a} &=& \frac{\partial F_t(S_t)}{\partial R_{t,a}},\label{eq:vhat}
\en

Once $\vhat^n_{t,a}$ is computed at iteration $n$, its value is used to update $\vbar^{n-1}_{t,a}$ giving us $\vbar^n_{t,a}$. The value function approximation is updated using
\bn
\vbar^n_{t,a} = (1-\alpha_{n})\vbar^{n-1}_{t,a} + \alpha_{n} \vhat^n_{t,a},\label{eq:smoothvhat}
\en
where $\alpha_{n}$ is a stepsize between 0 and 1.

\begin{figure}[tb]
{ {\footnotesize
\begin{description}
        \item[\textbf{Step 0:}] Initialization:
           \item[  ]\hspace{1cm} Set $n = 1$, initialize $\Vbar^0_t,~~t\in\Tcal$ and the state $S^{1}_0$.

        \item[\textbf{Step 1:}] \textbf{For }$n=1,\ldots,N$ \textbf{do}:
       \item[ \hspace{0.5cm} \textbf{Step 1a:}] Choose a sample path $\omega^n$.
        \begin{description}
           \item[ \hspace{1cm}\textbf{Step 2:}] For $t = 0, 1, \ldots, T$ do:
           \begin{description}
              \item[ \hspace{1cm}\textbf{Step 2a:}] Solve the optimization problem:
              \bn
              \max_{x_t\in\Xcal^n_t} \big(C_t(S^n_t,x_t) + \Vbar^{n-1}_t(S^{M,x}(S^n_t,x_t)) \big) \label{eq:findx}
              \en
              \hspace{0.7cm} Let $x^n_t$ be the value of $x_t$ that solves \eqref{eq:findx}, and $\vhat_{ta}$ be the dual corresponding to the resource conservation
              \item[ \hspace{1cm}\textbf{Step 2b:}] If $R_{ta_t} > 0$, update the value function using
              \bns
              \vbar^n_{t,a} = (1-\alpha_{n})\vbar^{n-1}_{t,a} + \alpha_{n}  \vhat^n_{t,a}
              \ens
              \item[ \hspace{1cm}\textbf{Step 2c:}]  Update the state:
              \bns
              S^{x,n}_t = S^{M,x}(S^n_t,x^n_t) \mbox{  and  } S^n_t    =& S^{M,W}(S^{x,n}_{t-1},W_t(\omega^n))
              \ens
           \end{description}
        \end{description}
        \item[\hspace{0.5cm}\textbf{}] \textbf{End For}
        \item[\textbf{Step 3:}] Return the value functions, $\{\vbar^n_{ta}, t=1,\dots,T, a \in \Acal\}$.
\end{description}
\caption{An approximate dynamic programming algorithm to solve the car assignment problem.}
\label{alg:adp}
}}
\end{figure}

The approximate dynamic programming algorithm that solves the fleet management problem is presented in Figure~\ref{alg:adp}. This algorithm employs a single pass to simulate a sample trajectory using the current estimates of the value functions.  It starts from an initial state $S^1_0 = (R_0,D_0)$ of cars and trips with a value function approximation $\Vbar_t^0(S^x_t)$.  Then, the assignment problem is solved to determine the actions of the cars to either take trips, reposition, recharge or stay, $x^1_0$; this is used to determine the post-decision state $S^{x,1}_0$ and then the next state $S_1 = S^{M,W}(S^{x,1}_0,W_1(\omega^1))$ is simulated.  This simulation includes both new trips, as well as changes to the status of the cars.  All of the complexity of the physics of the problem is captured in the transition functions, which impose virtually no limits on our ability to handle the realism of the problem.

A major technical challenge in the algorithm is computing the value function approximation $\Vbar_t = (\vbar_{ta})_{a\in\Acal}$.  Even if the attribute vector $a$ has only a few dimensions,  the attribute space is too large to be updated using \eqref{eq:smoothvhat}.  Furthermore, we only obtain updates $\vhat^n_{ta}$ for a subset of attributes $a$ at each iteration.
The proposed algorithm works perfectly well if each zone contains an observation, i.e., a car at each point in time from the beginning of the horizon till its end so that we can get an evaluation/measurment of the marginal value of a car. However, since the training is done on real data sets, then the observations are very sparse which does not capture the real value of a zone.

In principle, the decision problem can be solved for a resource vector $R_{t}$ that contains all attributes in $\Acal$ which is completely impractical. In the simulations, we only generated nodes for attributes $a$ where $R^n_{ta} >0$, which means that $\vhat^n_{ta}$ is only obtained for a subset of attributes; however, an estimate of $\vbar_{ta}$ is needed not just for where there are cars (that is, $R^n_{ta} >0$) but where cars might be sent. This problem is addressed in the next section.

\subsection{Hierarchical Aggregation for Approximating Value Functions}
\label{sec:hieragg}
We chose a lookup table representation for the value function $V_t(S_t)$ which is defined over a discrete state of states that captures time, location and battery charge level.  This was initially based on our poor experience using parametric models, and was then reinforced by empirical testing. To evaluate the value of these states, we need   observations of the function, $\vhat^n_{t,a}$ for every car attribute $a\in\mathcal{A}$, derived from simulations of the value of being in a state $S_t$, to create an estimate $\Vbar^{n+1}_t(S_t)$). The problem with lookup table representations is that if the variable $x$ (or state $S$) is a vector, then the number of possible values becomes exponentially larger with the number of dimensions. This is the classic curse of dimensionality.


One powerful strategy that makes it possible to extend lookup tables is the use of
hierarchical aggregation~(\citet{George2008}). Rather than simply aggregating a state space into a smaller space,
we pose a family of aggregations, and then combine these based on the statistics of our
estimates at each level of aggregation. This should not
be viewed as a method that ``solves'' the curse of dimensionality, but it does represent a
powerful addition to our toolbox of approximation strategies. In particular, we typically have to start with no data, and steadily build up an estimate of a function. We can accomplish this transition from little to no data, to increasing numbers of observations, by using hierarchical aggregation. Significantly, this method produces improved estimates as more observations become available, and offers asymptotic unbiasedness if we ultimately sample all zones infinitely often.

In this work, we simulated a network formed of $201\times 340$ zones, out which  $21643$ are valid zones. The battery levels are discretized into $20$ different levels (that is, from a charge level 0-5 percent up to 95-100 percent); this requires around $44000$ observations at each point in time to cover the full state space which is impractical because we do not need fleets of this size to begin with. Fortunately, it is not necessary to use all these attributes for the purpose of approximating the value functions.  Most of these attributes will never be visited, and many will be visited only a few times.  As a result, we have serious statistical issues in our ability to estimate~$\vbar_{ta}$.

\begin{figure}[t!]
\centering
\includegraphics[scale=0.5]{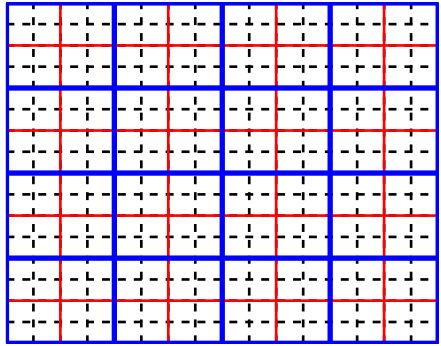}\vspace{-0.3cm}
\caption{\label{fig:aggregation} An example showing two aggregation levels of an area; the squares defined by the dotted lines correspond to the zones at the disaggregate level i.e., level $0$. The squares defined by the red and blue lines correspond to aggregation levels $1$ and $2$, respectively.}
\end{figure}

 We can use aggregation to create a hierarchy of state spaces, $\left\{\Acal^{(g)}, g = 0, 1, 2, \dots, |\Gcal|\right\}$, with successively fewer elements. In addition to time and battery level, we have a spatial correlation between neighboring zones; i.e., if two zones are close to each other then it is very likely that their value functions (over time and battery levels) have close values (e.g., two zones in urban or suburban areas). The aggregation structure that we are going to use in this work is spatial aggregation, that is at aggregation level $0$, we have $21643$ zones, at aggregation level $g>0$, we aggregate every $2^{2*g}$ neighboring zones together into an area. If we have an observation (a car with attribute $a_t$) in a certain zone, this gives an estimate of the value function at $a_t$ for all zones that are aggregated with it.   For every aggregation level, a zone belongs to one and only one of the areas. We illustrate five levels of aggregations in Table~\ref{tab:nomadicaggregations} and two level of aggregation in Figure~\ref{fig:aggregation}.

We are going to estimate a function at different levels of aggregation. We can assume that the estimate of the function at the most disaggregate
level is noisy but unbiased, and then let the difference between the function at some level of aggregation and the function at the most disaggregate level as an estimate of the bias.

\begin{table} [tb]
\begin{center}
\begin{tabular}{|c|c|c|c|c|c|} \hline
Agg. level $g$      & 0 & 1 & 2 & 3 & 4\\ \hline \hline
Zones/level      & 1 & 4 & 16 & 64 & 256\\ \hline \hline
Areas & 21643  &  7051 & 2447 & 950 & 326   \\ \hline
\end{tabular}
\vspace{0.3cm}
\end{center}\vspace{-0.0cm}
\caption{Levels of aggregation used to approximate value functions. The aggregation is done on the zones where every $2^{2\cdot g}$ zones form an area. All areas are disjoint at each aggregation level. }
\label{tab:nomadicaggregations}
\end{table}

Choosing the right level of aggregation to approximate the value functions involves a tradeoff between statistical and structural errors. If $\left\{\vbar^{(g)}_{ta},{g\in\Gcal}\right\}$ denotes estimates of a value $\vbar_{ta}$ at different levels of aggregation, we can compute an improved estimate as a weighted combination of estimates of the values at different levels of aggregation using
\bn
\vbar_{ta} = \sum_{g \in \Gcal} w^{(g)}_{ta} \cdot \vbar^{(g)}_{ta},  \label{eq:weightedaverage}
\en
where $\{w^{(g)}_{ta}\}_{g \in \Gcal}$ is a set of appropriately chosen weights. \cite{George2008} shows that good results can be achieved using a simple formula, called WIMSE, that weights the estimates at different levels of aggregation by the inverse of the estimates of their mean squared deviations (obtained as the sum of the variances and the biases) from the true value.  These weights are easily computed from a series of simple calculations; here we briefly summarize the equations without deriving them.  We first compute
\bn
\betabar^{(g,n)}_{ta}       &=& \textwrap{Estimate of bias due to smoothing a transient data series,}\nonumber\\
                            &=& (1-\steptwo_{n-1})\betabar^{(g,n-1)}_{ta} + \steptwo_{n-1} (\vhat^n_{ta} - \vbar^{(g,n-1)}_{ta})\label{eq:betabar}.\\
\mubar^{(g,n)}_{ta}         &=& \textwrap{Estimate of bias due to aggregation error,}\nonumber\\
                            &=& \vbar^{(g,n)}_{ta} - \vbar^{(0,n)}_{ta}. \nonumber\\
\bar{\bar{\beta}}^{(g,n)}_{ta} &=& \textwrap{Estimate of total squared variation,} \nonumber\\
                            &=& (1-\steptwo_{n-1}) \bar{\bar{\beta}}^{(g,n-1)}_{ta} + \steptwo_{n-1} (\vhat^n_{ta} - \vbar^{(g,n-1)}_{ta})^2.\nonumber
\en
We are using two stepsize formulas here.  $\step^{(g,n-1)}_{ta}$ is the step size used in equation \eqref{eq:smoothvhat} to update $\vbar^{n-1}_{ta}$. $\steptwo_n$ is typically a deterministic step size which might be a constant such as 0.1.

We estimate the variance of the observations at a particular level of aggregation using
\bn
(s^2_{ta})^{(g,n)}=\frac{\bar{\bar{\beta}}^{(g,n)}_{ta} - (\betabar^{(g,n)}_{ta})^2}{1+\lambda^{(g,n)}_{ta}},  \label{eq:ssquared}
\en
where $\lambda^{(g,n)}_{ta}$ is computed using
\bns
\lambda^{(g,n)}_{ta} = \begin{cases} (\step^{(g,n-1)}_{ta})^2 & \text{$n=1$},\\
                      (1-\step^{(g,n-1)}_{ta})^2\lambda^{(g,n-1)}_{ta} + (\step^{(g,n-1)}_{ta})^2 & \text{$n>1$}. \end{cases}
\ens
This allows us to compute an estimate of the variance of $\vbar^{(g,n)}_{ta}$ using
\bn
(\sigmabar^2_{ta})^{(g,n)} &=& \Var[\vbar^{(g,n)}_{ta}] = \lambda^{(g,n)}_{ta} (s^2_{ta})^{(g,n)}. \label{eq:sigmabar}
\en
The weight to be used at each level of aggregation is given by
\bn
w^{(g,n)}_{ta} \propto \left((\sigmabar^2_{ta})^{(g,n)} + \left(\mubar^{(g,n)}_{ta}\right)^2\right)^{-1},
%
\en
where the weights are normalized so they sum to one.
This formula is easy to compute even for very large scale applications such as this.  All the statistics have to be computed for each attribute $a$, for all levels of aggregation, that is actually visited.  From this, we can compute an estimate of the value of any attribute regardless of whether we visited it or not.

\subsection{Monotonicity of the Value Functions}\label{sec:Montone_VFs}
We use a value function approximation that is linear in the resource vector $R_t$ as in (\citet{Simao2009}); however, in this work, we exploit two additional structural properties to accelerate learning:
\begin{itemize}
\item[1)] The value function is monotone increasing in the charge level of the battery.
\item[2)] The value function is monotone decreasing with time.
\end{itemize}
Our use of these structural properties proves to be critical to the success of the methodology. This is the first time we use ADP with monotone value functions for fleet management.
Suppose that the attribute space $\mathcal{A}$ is equipped with a partial order, denoted $\preceq$, and the following monotonicity
property is satisfied for every $t$:\vspace{-0.5cm}
\bn
a\preceq a' \Rightarrow v^*_{ta}\leq v^*_{ta'}. \label{eq:monotone_battery}
\en
This means that the optimal marginal value function $v^*_{ta}$ is order preserving over the attribute space $\mathcal{A}$. An example that arises very often is the following definition of $\preceq$, which we call the generalized componentwise inequality. Each attribute $a$ can be decomposed into  $a=(a_1,a_2)$ for $a_1\in\mathcal{Z}$, $a_2\in \mathcal{L}$. For two attributes $a=(a_1,a_2)$ and $a=(a_1',a_2')$, we have
\bn
a\preceq a' \Rightarrow  a_1=a_1', a_2\leq a_2'.
\en
This means that if we hold $a_1$ constant, then the value function is monotone in the second variable $a_2$. This applies to our model since a higher battery means a higher value for the car as it can take more and longer trips.

\begin{proposition}\label{app:proposition1_def}
Let $a_t=(a_{t1},a_{t2})$ be an attribute vector of a car at time $t$. Let $\preceq$ be the partial order on $\mathcal{A}$ as described in \eqref{eq:monotone_battery}. Assume the following assumptions hold.
\begin{itemize}
\item[(i)] For every $a_t,a_t'\in\mathcal{A}$ with $a_t\preceq a_t'$, $d\in \mathcal{D}$ and $w\in\mathcal{W}_t$, the transition attribute function satisfies\vspace{-0.3cm} \bns a^{M}(a_t,d)\preceq a^{M}(a_t',d) \mbox{  and  }\ a^{M,W}(a_t,d,w)\preceq a^{M,W}(a_t',d,w).\ens \vspace{-0.9cm}
\item[(ii)] For each $t\leq T$, $a_t,a_t'\in\mathcal{A}$ with $a_t\preceq a_t'$ and $d\in\mathcal{D}$, $c_{tad}\leq c_{ta'd}$ and $c_{Ta}\leq c_{Ta'}$.
\item[(iii)] For each $t<T$, $a_{t2}$ and $W_{t+1}$ are independent.
\end{itemize}
Then, the pre-decision and post-decision value functions $v^*_{t}$ are monotonically increasing in the battery dimension.
\end{proposition}
\emph{Proof }. See the appendix.


\begin{proposition}\label{app:proposition2_def}
Let $a=(a_{1},a_{2})$ be an attribute vector of a car. The value function $v^*_{t}$ is monotically decreasing with time for any $a\in\mathcal{A}$, i.e.,
\bn
t\leq t' \Rightarrow v^*_{ta}\geq v^*_{t'a}. \label{eq:monotone_time}
\en
\end{proposition}
\emph{Proof }. See the appendix. This can be justified by the fact that the value of a vehicle with the same attribute declines with time since there is less time to earn money.

The observation of the value function at time $t$, iteration $n$, and with attributes $a_t$ is denoted by $\vhat^n_{ta}$ and is calculated using the estimate of the value function from iteration $n-1$. This observation is then smoothed with $\vbar^{n-1}_{ta}$ to get an updated value function, $\vbar^{n}_{ta}$, using a stochastic approximation step.  Let $\Pi_B$ be the monotonicity preserving projection operator in the battery dimension of the value function. Note that the term ``projection" is being used loosely here; the space that we ``project" onto actually changes with each iteration.

\begin{definition}
For $a'\in\mathcal{A}$ and $h_{ta'}\in\mathbb{R}$, let $(a',h_{ta'})$ be the reference point to which other states are compared. Define the projection operator $\Pi_B: \mathcal{S}\times \mathbb{R}\times\mathbb{R}^2\rightarrow\mathbb{R}$ where the component of the vector $\Pi_B(a',h_{ta'},v_{ta})$ at attribute $a$ is given by
\bns
\Pi_B(a',h_{ta'},v_{ta})= \left\{
\begin{array}{ll}
h_{ta'}, & \mbox{if $a = a'$}, \\
h_{ta'}\vee v_{ta}, & \mbox{if $a'\preceq a, a\neq a'$}, \\
h_{ta'}\wedge v_{ta}, & \mbox{if $a'\succeq a, a\neq a'$}, \\
v_{ta}, & \mbox{otherwise}.
\end{array}
\right.
\ens
\end{definition}
where $v_{ta}$ is the current value function approximation for attribute $a$, $(a',h_{ta'})$ is the smoothed observation for attribute $a'$ at time $t$, and $\Pi_B(a',h_{ta'},v_{ta})$ is the updated value function approximation. Violations of the monotonicity property of \eqref{eq:monotone_battery} are corrected by $\Pi_B$ in the following ways:
\begin{itemize}
\item if $h_{ta'}\geq v_{ta}$ and $a'\preceq a$, then $v_{ta}$ is too small and is
increased to $h_{ta'} = h_{ta'}\vee v_{ta}$ and
\item if $h_{ta'}\leq v_{ta}$ and $a'\succeq a$, then $v_{ta}$ is too large and is
decreased to $h_{ta'} = h_{ta'}\wedge v_{ta}$.
\end{itemize}
\citet{Jiang2015} have shown that choosing the latest observed value for satisfying the monotonicity constraint is asymptotically optimal.
\begin{figure}[t!]
\centering
\includegraphics[scale=0.58]{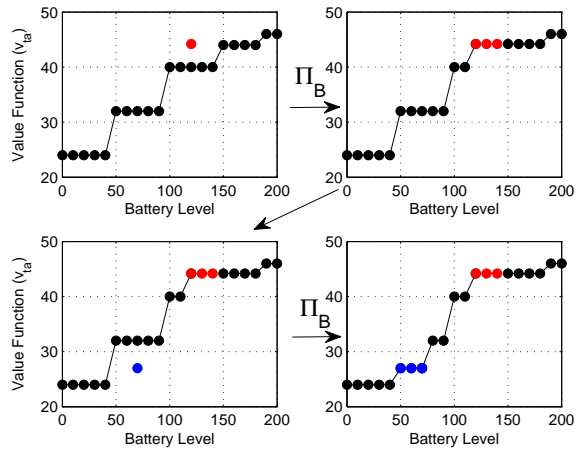}
\caption{\label{fig:monotone} An example illustrating the projection operator $\Pi_B$.}
\end{figure}
Figure~\ref{fig:monotone} shows an example of a sequence of two observations for the battery level, specified by attribute $a_2$, and the projection operator for the value function $v_{ta}$.

Similarly, we define $\Pi_T$ as the monotonicity preserving projection operator in the time dimension of the value function. Let $h_{t'a}$ be the latest observation of the smoothed value function at time $t'$ for attribute $a$.
\begin{definition}
For $a\in\mathcal{A}$ and $h_{t'a}\in\mathbb{R}$, let $(a,h_{t'a})$ be the reference point to which other states are compared. Define the projection operator $\Pi_T: \mathcal{S}\times \mathbb{R}\times\mathbb{R}^2\rightarrow\mathbb{R}$ where the component of the vector $\Pi_T(a,h_{t'a},v_{ta})$ at attribute $a$ is given by
\bns
\Pi_T(a,h_{t'a},v_{ta})= \left\{
\begin{array}{ll}
h_{t'a}, & \mbox{if $t = t'$}, \\
h_{t'a}\vee v_{ta}, & \mbox{if $t' > t$}, \\
h_{t'a}\wedge v_{ta}, & \mbox{if $t'< t$}, \\
v_{ta}, & \mbox{otherwise}.
\end{array}
\right.
\ens
\end{definition}

\subsection{Discussion on Convergence}
Solving this problem exactly using classical methods for discrete dynamic programming is computationally intractable as we have shown in Section~\ref{sec:lookahead_policy}; hence, approximations are crucial to get a scalable policy. Basic lookup table approximations worked very poorly, and we quickly realized that parametric approximations simply would not capture the complex structure of the value function.  We adopted two strategies that proved critical to the success of the method. First, we used hierarchical aggregation to get better estimates of the value functions with a small number of observations. We were also able to prove that the value functions were monotone in both time (the earlier in the horizon the larger the value) and battery charge level (the higher the charge, the higher the value); using the monotonicity property accelerates convergence.

There is some theoretical work to support the use of both of these strategies. First, \citet{Jiang2015} show that exploiting monotonicity of the value function is asymptotically optimal and experimentally increases the rate of convergence by around two orders of magnitude. This proof was done in the context of approximate dynamic programming, but without using hierarchical~aggregation.

We have had prior experience using hierarchical aggregation in fleet management, but as of this writing, we do not have a formal convergence proof.  \citet{George} derive optimal weights for the hierarchical aggregation originally developed in the context of stochastic search  but do not prove that the estimates produced are asymptotically unbiased.  This proof is given in (\citet{Mes2011}) when  using a learning policy that ensures that all points are visited infinitely often, but this is not in the context of approximating value functions which is left as a future work. Hierarchical aggregation can have significant performance gains when the number of observations is small; however, if we have enough observations in the system then hierarchical aggregation is not needed; this depends on the size of the system and the available data~set.

The only approximation used to solve this problem is that the value function is linear with the number of cars. If there is only one car in the system, the linear approximation holds and the proposed policy for the dispatch problem is asymptotically optimal. For a fleet with a small number of cars relative to the market it is serving, it is logical to assume that the value of additional cars is linear; that is, if every car is gaining an amount of $x$ dollars, the added additional car will also gain on average $x$ dollars since all cars are operating all the time. However, when adding a large number of cars in the system, not all of them will be operating all the time and the marginal value of an added car decreases. An alternative strategy would be to use piecewise linear, separable approximations as suggested in (\citet{Godfrey2002}) and (\citet{Topaloglu2006}) but this induces other computational challenges that affect~performance.

We conclude that we have an asymptotically optimal policy for modeling a single car; however, this does not apply for larger fleet sizes. We have proposed an optimized policy that is scalable and tried to induce minimal assumptions (linear value functions) for solving very large realistic models. We have also exploited techniques such as monotonicity to increase the convergence rate of the policy and hierarchical aggregation to get better estimates of the value function with a small number of observations.

\section{The Surge Pricing Problem}\label{sec:Surge_Pricing}
The success of ridesharing fleets requires the use of what Uber refers to as ``surge pricing'' to balance the supply of cars and riders.
The objective of surge pricing is maximizing the revenue by maximizing the offered prices by the operator while still being accepted by the riders.  When a rider requests a trip, the operator offers a price that might be either accepted or rejected by the rider. Hence, the operator must learn the acceptance rate of the customer based on the offered price. The operator must also learn its acceptance rate for an offered price; for example, if the offered price is too low, then it is better for the operator to reject the low price and hold the car for a better offered price. The fleet operator needs to learn two models: 1) the probability that a rider will accept a ride at a price, and 2) the probability that the operator will allocate a car to a rider at that price.  We assume that there is a learner who is trying to learn the acceptance rates of both the operator and the rider.

In this work, we assume parametric response curves for the operator and rider with parameters for each zone, producing
a high dimensional parametric model that needs to be estimated based on the responses from each transaction. We use discrete choice models (such as the logit family of supply/demand functions) for both the operator and the rider. We can maximize the revenue if we know the true response curves in each zone, but this information is simply not available. To this end, we adopt a Bayesian approach and use a prior distribution over model parameters to express our uncertainty about the operator/rider responses. We present the mathematical formulation of the problem in Section~\ref{sec:math_model_surge_pricing} and then present a resampling approach for the response curves in Section~\ref{sec:bootstrap_aggregation}.

\subsection{The Mathematical Model}\label{sec:math_model_surge_pricing}
At each decision epoch $t$ in the ADP algorithm presented in Figure~\ref{alg:adp}, we collect all trips, i.e., we find $D_t$.  We assume that there is one price per zone (expressed in dollars per mile) from where the trips originate at each point in time which is specified by the attribute vector of the trip $b$.  To find this price, we need a belief state about the distribution of the prices in every zone denoted by $B_t$ in Section~\ref{sec:Prob_Formulation}. In what follows, we explain how we calculate and use the belief state to find the price for each trip.

For an incoming trip, the learner is presented with a set of attributes $b$ and it has to decide on the price $p^m_b$ as denoted in Equation~\eqref{eq:rewardfunction}. We assume that we have a discretized set $\tilde{P}$ of possible prices from a given interval $[\tilde{p}^{\mbox{min}},\tilde{p}^{\mbox{max}}]$.

The originating zone, $z$, of a trip with attribute, $b_{t}$, is specified by the trip's first attribute, i.e., $z=b_{t,1}$.  For any attribute and price, there is an unknown underlying binomial probability of operator and rider acceptance,  designated by $f_z^o(b, \tilde{p};\gamma_z)$ and $f_z^r(b, \tilde{p};\beta_z)$ as the probability of responding ``accept."  For each zone $z$, we use two logistic models, one for operator $f_z^o(b_t, \tilde{p}_t;\gamma_z)= \sigma(\gamma_z^T\cdot x^o(b_t, \tilde{p}_t))$, and one for rider $f_z^r(b_t, \tilde{p}_t;\beta_z)= \sigma(\beta_z^T \cdot x^r(b_t, \tilde{p}_t))$, where $\sigma(h)=\frac{1}{1+e^{-h}}$ , $x^o(b_t,\tilde{p}_t)$ and $x^r(b_t, \tilde{p}_t)$ are the covariates constructed from trip attributes and the price.

In this work, we assume that $x^o(b_t, \tilde{p}_t) = [1,t,\tilde{p}_t,N_t^v(b_{t,1}),N_t^r(b_{t,1})]^T$ where $b_{t,1}$ is the originating zone of the trip, $\tilde{p}$ is the price, $N_t^v(b_{t,1})$ and $N_t^r(b_{t,1})$ are the number of vehicles and trips, respectively at time $t$ in zone $b_{t,1}$. This means that vector $\gamma$ is five-dimensional i.e., $\gamma=[\gamma_0,\gamma_1, \gamma_2,\gamma_3,\gamma_4]^T$. The rider has less features in its model given by $x^r(b_t, \tilde{p}_t) = [1,t,\tilde{p}_t]^T$ since his acceptance probability depends on the time of the day and the offered price; hence, the vector $\beta$ is three dimensional.

\begin{figure}[t!]
\centering
\vspace{-0.5cm}
\subfigure[]{
\includegraphics[scale = 0.5]{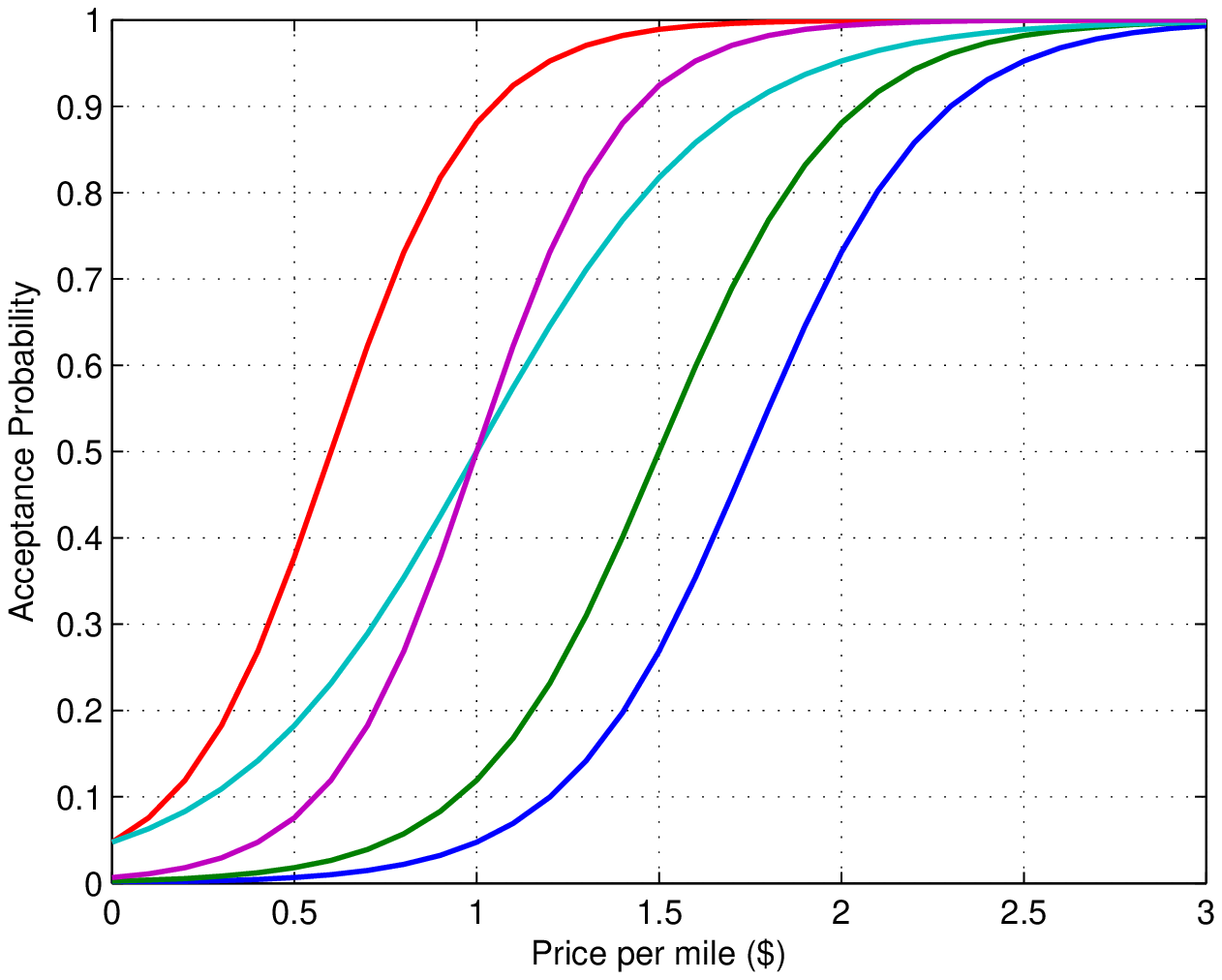}
\vspace{-0.0cm}
\label{fig:logistic_supply}
}
\vspace{-0.0cm}
\subfigure[]{
\includegraphics[scale = 0.5]{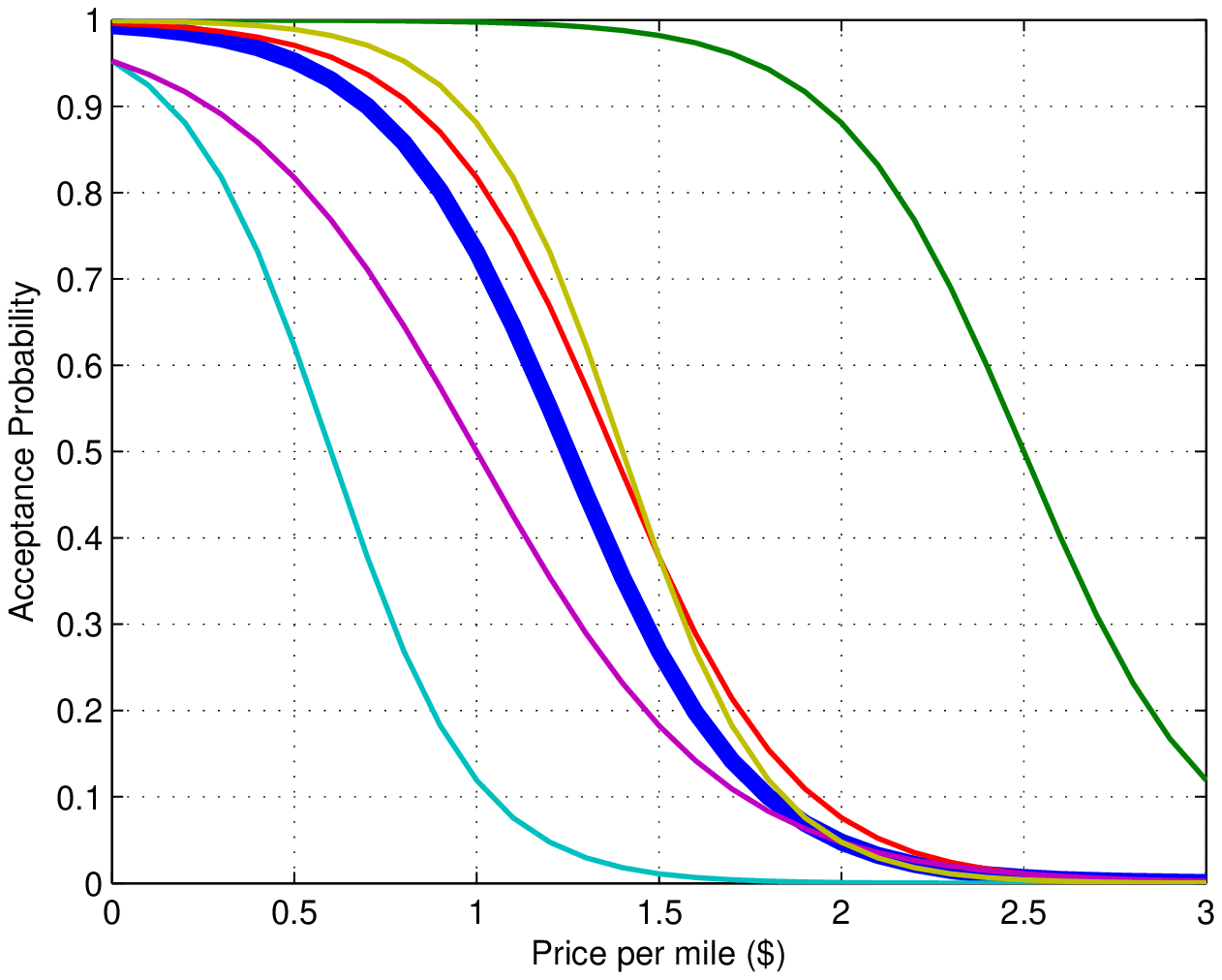}
\vspace{-0.0cm}
\label{fig:logistic_demand}
}
\caption{\label{fig:acceptance_model} Operator and Rider acceptance models generated randomly for one zone assuming $K=25$ curves, i.e., $5$ for the operator and $5$ for the rider. (a) Operator model. (b) Rider model which shows $5$ random curves and the true curve for the zone (blue one).}
\end{figure}

Following the work by \citet{Harrison2012} and \citet{ChenReyes2015}, we begin with a sampled
belief model where the learner adopts $K$ ambient demand and supply models, for each zone, from the known logistic family $f_{z,k}(b_t,\tilde{p}_t) := f_z(b_t, \tilde{p}_t;\theta_{z,k})$.  We assume that there is a learner who is supposed to learn which one of the $K$ curves is the true one for each zone. Figure~\ref{fig:acceptance_model} presents an example of operator and rider logistic curves for a particular zone. It shows $K=25$ random curves which corresponds to $5$ logistic curves for the operator and $5$ curves for the rider, respectively. We also assume that we have a true curve for the rider which mimics how riders decide in real life.

The learner uses the $K$ curves for choosing a price $\tilde{p}_t$ for each trip at decision epoch $t$ as will be discussed later. Based on the offered price, the rider will accept or reject with a certain probability as happens in real life and we are able to collect the rider's response $y_t^r(b_t,\tilde{p}_t)$ through an application on their smartphone. Hence,  in real life, there is a true response curve for the riders in every zone denoted by $f_z^{r,\mbox{\small true}}(b_t,\tilde{ p}_t)$ that the learner must learn.

In our simulations, we assume that every zone has a true response curve for the riders; however it is not revealed to the learner.  The response of the rider, $y_t^r$, is based on a Bernoulli random variable with probability of success   $f_z^{r,\mbox{\small true}}(b_t,\tilde{ p}_t)$; if $y_t^r=1$, it means that the rider accepts the offered price and the operator has to assign it to one of the cars (if possible). At decision epoch $t$, we collect all trips whose offered prices have been accepted by the riders and assign them to cars by solving the assignment problem in Step 2a) of the ADP algorithm. Based on the solution of the assignment problem, we get a response from the operator $y_t^o$ for each trip which determines whether it was assigned a car, i.e., $y_t^o=1$, or not. For each trip, the learner will use the two binary responses from the operator and rider $y_{t} = (y^o_t,y^r_t) \in\{-1,1\}$. If the offered price is accepted by both the operator and the rider, the learner can get a revenue proportional to the price $\tilde{p}$; otherwise, the revenue is~zero.


The learner's challenge is to offer prices to maximize his total revenue over time $\sum_{t=0}^T \tilde{p}_t \cdot Y^o_t(b_t, \tilde{p}_t)Y^r_t(b_t, \tilde{p}_t)$. It can be generally assumed that $Y^o$ and $Y^r$ are independent, meaning that whether the operator will accept a trip has nothing to do with whether a rider will accept the trip at the same time. Hence, we define the learner's single-step utility function which calculates the expected value of a contract based on its price and the likelihood it will be accepted by both the operator and the rider:
\bn
f_{z,k}(b_t,\tilde{p}_t;\theta_{z,k}) = \mathbb{E}[\tilde{p}_t\cdot Y_t^oY_t^r] = \tilde{p}_t\cdot f_{z,k}^o(b_t, \tilde{p}_t;\gamma_{z,k})\cdot f_{z,k}^r(b_t,\tilde{p}_t;\beta_{z,k}).
\en

 For each zone, we denote by $q_{0,z,k}$ the multinomial prior probability of the unknown parameter values $\theta_{z,k}$ at time $t=0$, assigned by the learner, to each candidate $f_{z,k}(b_t,\tilde{p}_t;\theta)$ and $q_{t,z,k}=p(\theta_z=\theta_{z,k}|\mathcal{O}_{tz})$ as the posterior probability of the $k^{\mbox{th}}$ candidate given the previously observed data points up to time $t$, $\mathcal{O}_{t,z} = \left\{\left(\tilde{p}_i,(y^o_i,y^r_i)\right)\right\}_{i=0}^t$. After we make the next decision $\tilde{p}_{t+1}$ under current trip attributes $b_{t+1}$ and obtain the next observation $y_{t+1}$, by Bayes' rule, we have
\bn
q_{t+1,z,k} &=& p(\theta_z=\theta_{z,k}|\mathcal{O}_{t+1,z})\nonumber\\
          &\propto& p(y_{t+1}|\theta_z=\theta_{z,k},\mathcal{O}_{t,z})p(\theta_z = \theta_{z,k}|\mathcal{O}_{t,z})\nonumber\\
          &=& q_{t,z,k} p(y^o_{t+1}|\gamma_z = \gamma_{z,k})p(y^r_{t+1}|\beta_z = \beta_{z,k})\nonumber\\
          &=& q_{t,z,k} \sigma(y^o_{t+1}\cdot\gamma^T_{z,k} \cdot x^o(b_{t+1}, \tilde{p}_{t+1}))\sigma(y^r_{t+1}\cdot\beta^T_{z,k} \cdot x^r(b_{t+1}, \tilde{p}_{t+1}))\label{eq:post_prob}
\en
For every zone with location $z$, we define the belief state at time $t$ to be the probability distribution of the prices in that zone given by:
\bn
B_{tz} = \left(q_{t,z,1},\ldots,q_{t,z,K} \right)
\en
Now, we can define the belief state for all zones as $B_t=\{B_{tz}\}_{z\in\mathcal{Z}}$ which is used to determine the price for a trip request.

Using the belief state, we adopt a pure exploitation policy that chooses a price that maximizes the revenue, considering the response curves for both the fleet operator and the rider. This means that for each trip request, $b_t$, we solve
\bn
\tilde{p}^*_t = \arg\max_{\tilde{p}_t}\tilde{p}_t\sum_{k=1}^Kf_z(b_t,\tilde{p}_t|\theta_{z,k})q_{t,z,k},\mbox{~~where~~}z=b_{t,1}. \label{eq:opt_price}
\en
Note that Equation~\eqref{eq:opt_price} yields the same price for all trips originating from the same zone. After recommending a price for each trip, we collect the responses from the riders and then call the assignment problem in Step 2a) of the ADP algorithm to get the responses from the operator. Then, the belief state is updated based on the obtained responses; all these steps are summarized in the surge pricing algorithm presented in Figure~\ref{alg:surge_pricing}.

\subsection{Bootstrap Aggregation}\label{sec:bootstrap_aggregation}
We do not know the true parameter value $\theta_z$ so we need to gather information to estimate it.  Starting with a random set of logistic curves and assuming that one of the candidate values of $\theta$ is the true one is too naive. This may be a good approximation if $\theta_z$ is a one or two
dimensional vector and we generate a large set of curves. On the other hand, after several experiments, it is likely that the probability
mass is concentrated on a small subset of the original candidates, indicating that others are less
likely to be the true model. To this end, we propose a bootstrap aggregation method to constantly
resample the candidate set so that it can adaptively find more promising parameter~values.

If resampling is triggered at time $t$, meaning that we re-generate the candidate set of $K$ parameter
values aiming to include more promising values, we draw on a technique known as bagging~(\citet{Breiman1996}) which works as follows. At time $t$, we already have a set of $m$ observations of prices and responses denoted by $\mathcal{O}_{t,z}$ for each zone $z$. For each zone, we generate $K$ new training data sets $\mathcal{O}_{t,z,k}, k=1,\ldots,K$, by sampling from $\mathcal{O}_{t,z}$ uniformly at random and with replacement. Then $K$ models are fitted by using logistic regression and finding the maximum likelihood estimator $\theta_{t,z,k}$, based on the above $K$ bootstrapped samples.
These $K$ models are fixed as the ambient candidates until the next resampling step. In traditional
bagging, the $K$ models are combined by averaging the output. Instead, we treat the equal probability
as the initial prior distribution of the $K$ bootstrapped models. We then make use of previously
observed data $\mathcal{O}_{t,z,k}$ and obtain the posterior distribution by calculating the likelihood of each model,
\bn
\mathcal{L}(\theta_{t,z,k}|\mathcal{O}_{t,z}) = \prod_{i=0}^t  \sigma(y^o_{i}\cdot\gamma^T_{t,z,k} \cdot x^o(b_{i}, \tilde{p}_{i}))\sigma(y^r_{i}\cdot\beta^T_{t,z,k} \cdot x^r(b_{i}, \tilde{p}_{i}))
\en
With resampling, we modify the posterior probabilities as
\bn
q_{t,z,k}=\frac{\mathcal{L}(\theta_{t,z,k}|\mathcal{O}_{t,z,k})}{\sum_{i=1}^K\mathcal{L}(\theta_{t,z,i}|\mathcal{O}_{t,z,i})}\label{eq:resample}
\en
The resampling criteria can be set to different conditions. In this work, we sample by the end of the time horizon of each iteration of the ADP algorithm. Figure~\ref{alg:surge_pricing} summarizes all the steps that we have discussed for surge pricing.

\begin{figure}[tb]
{ {\footnotesize
\begin{description}
        \item[\textbf{Step 0:}] Initialization: initialize $\theta_{z,k}$ and $q_{0,z,k}=1/K, k=1,\ldots,K$; initialize prices $\tilde{p}_1,\ldots,\tilde{p}_M\in[\tilde{p}^{\mbox{min}},\tilde{p}^{\mbox{max}}]$.

        \item[\textbf{Step 1:}] At each decision epoch $t$ in the ADP algorithm presented in Figure~\ref{alg:adp} do:
       \item[ \hspace{0.5cm} \textbf{Step 2:}] Collect all trips between decision epochs $t-1$ and $t$, i.e., find $D_t$.
        \begin{description}
           \item[ \hspace{1cm}\textbf{Step 2a:}] Use the belief state $B_t$ to determine the price for each trip in $D_t$; that is, for each trip with attributes
               \item[ \hspace{2.5cm}] $b_t,$ recommend a price $\tilde{p}^*_t $ by solving
           \bns
           \tilde{p}^*_t = \arg\max_{\tilde{p}_t}\tilde{p}_t\sum_{k=1}^Kf_z(b_t,\tilde{p}_t|\theta_{z,k})q_{t,z,k},\mbox{~~where~~}z=b_{t,1}.
           \ens
           \item[\hspace{1cm}\textbf{Step 2b:}] For each trip, get response $y_t^r$ from the rider based on bernoulli random variable ($\mathcal{B}(f_z^{r,\mbox{\small true}}(b_t,\tilde{p}_t))$.
               \item[\hspace{1cm}\textbf{Step 2c:}] After collecting all responses, solve the assignment problem in Step 2a) of the ADP algorithm.
              \item[\hspace{1cm}\textbf{Step 2d:}] Get responses, $y_t^o$, from the operator (based on Step 2b).
              \item[\hspace{1cm}\textbf{Step 2e:}] Update the posterior probabilities according to \eqref{eq:post_prob}
              \bns
               q_{t+1,z,k} \propto p(y_{t+1}|\theta_z=\theta_{z,k},\mathcal{O}_{t,z})p(\theta_z = \theta_{z,k}|\mathcal{O}_{t,z})
              \ens
               \item[\hspace{1cm}\textbf{Step 2f:}] Update the belief state $\mathcal{B}_{t+1,z} = \{q_{t+1,z,1},\ldots,q_{t+1,z,K}\}$ and  $\mathcal{B}_{t+1}=\{B_{t,z}\}_{z\in\mathcal{Z}}$.
              \item[\hspace{1cm}\textbf{Step 2g:}] Update the set of observations, $\mathcal{O}_{t+1,z} = \mathcal{O}_{t,z}\cup \left\{\tilde{p}_t,y_t\right\}$.
        \end{description}
        \item[ \hspace{0.5cm} \textbf{Step 3:}] if $t = T$:
        \begin{description}
              \item[\hspace{1cm}\textbf{Step 3a:}] Get responses the bootstrapped model, $\{\theta_{z,k}\}_{k=1}^L$, from $\mathcal{O}_{t,z},\forall z\in\mathcal{Z}$.
              \item[\hspace{1cm}\textbf{Step 3b:}] Set the posterior probabilities $q_{t,z,k}=\frac{\mathcal{L}(\theta_{t,z,k}|\mathcal{O}_{t,z,k})}{\sum_{i=1}^K\mathcal{L}(\theta_{t,z,i}|\mathcal{O}_{t,z,i})}$ according to \eqref{eq:resample}.
              \end{description}

\end{description}
\vspace{0.3cm}
\caption{The surge pricing algorithm.}
\label{alg:surge_pricing}
}}
\end{figure}

\section{The Fleet Size Problem}\label{sec:FleetSize_Problem}
The profit of the fleet is the revenue minus the cost of the cars when operating for $y^c$ years. Assume that the car operated for $d^o$ days per year. To account for the capital costs of the car, we need the following parameters. The base cost of the car without battery, $c^{\mbox{\footnotesize car}}$, the maintenance and insurance cost per year, $c^{\mbox{\footnotesize maint}}$, and the battery cost, $c^{\mbox{\footnotesize bat}}(b^{\mbox{\footnotesize size}})$, which increases non-linearly with the battery size~$b^{\mbox{\footnotesize size}}$. Now, we can define the profit of the fleet for $N^{\mbox{\footnotesize cars}}$ cars with battery size $b^{\mbox{\footnotesize size}}$ as:
\bn
\mbox{Profit}(N^{\mbox{\footnotesize cars}},b^{\mbox{\footnotesize size}}) = F_0(S_0) \cdot y^c \cdot d^o - N^{\mbox{\footnotesize cars}}(c^{\mbox{\footnotesize car}} + c^{\mbox{\footnotesize maint}} \cdot y^c + c^{\mbox{\footnotesize bat}}(b^{\mbox{\footnotesize size}})). \label{eq:economics}
\en
where $F_0(S_0)$ is the total revenue per day for a given fleet size  $N^{\mbox{\footnotesize cars}}$ with battery size $b^{\mbox{\footnotesize size}}$ which are incorporated in the initial state $S_0$. To find the optimal fleet size, we have to use the results of the dispatch and surge pricing problems to guide the design of the fleet size. 

We assume that all the cars have the same battery capacity.  An interesting topic for future research is to design a search algorithm to identify the mix of battery capacities to see if this is more effective.


\section{Numerical Results}\label{sec:results}
The policies are implemented in a simulator to address several questions:  1) How effective is the VFA policy for guiding efficient dispatch decisions compared to the myopic policy? 2) Will the VFA policy produce realistic, effective repositioning and recharging behaviors (such as avoiding recharging during peaks)?  3) How much does hierarchical aggregation improve performance? 4) What is the value of using surge pricing?  5) What is the tradeoff between fleet size and battery capacity (for example, if we use smaller fleets do we need larger batteries?).


\subsection{Data and Model Parameters}
\begin{figure}[t!]
\centering
\includegraphics[scale=0.5]{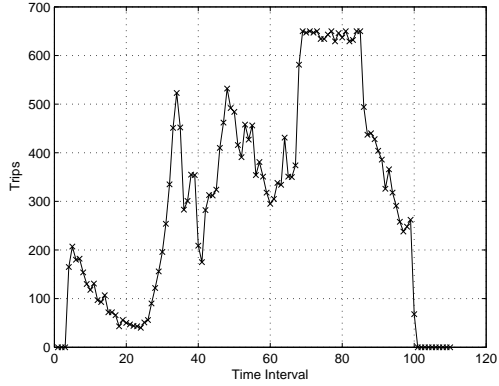}
\caption{\label{fig:trip_dist} Trip distribution over 24 hours divided into 15 minute increments.}
\end{figure}

To assess the performance of the proposed policies, the simulated ride-sharing system is constructed using real data of trips for the state of New Jersey available online (see \citet{TripData}). Figure~\ref{fig:trip_dist} shows the trip distribution over 24 hours with peak periods in the morning, mid-day (around lunch), and late afternoon which are typical of transportation patterns. The total number of trips is $32874$ divided into $15$ minute time increments which results in $96$ time periods; three time periods are added before the beginning of the day to reposition cars and $11$ time periods are added at the end so that all trips end in the simulation horizon resulting in 110 time periods. The $5^{\mbox{th}}$ percentile, $95^{\mbox{th}}$ percentile and average trip distances are $6$, $57.5$ and $24.8$ miles, respectively.  The minimum number of trips in a period is $40$ whereas the maximum is $650$. In the simulations, we sample from the given trip distribution.  The state of New Jersey is divided into $201$ by $304$ rectangular zones of width $0.5$ miles with $21634$ valid zones.


There are various simulation parameters that have a direct effect on the generated revenue. First, we analyze the various policies for a particular set of parameters and then we  present a study of the effect of the various simulation parameters. The simulation parameters used are presented in Table~\ref{tab:sim_params}. Note that every kWh of the battery level enables the car to travel for around $3$ miles, so we present the battery level as a function of either (kWh) or miles. We run the initial tests with a fleet of $1500$ cars, which provides a $95$ percent trip coverage.  Later,  the fleet is varied to study the economics of different fleet sizes, battery capacity and other operating parameters.

\begin{table} [tb]
\begin{center}
\begin{tabular}{|l|c|l|c|} \hline
Trip base fee - $\rho^m $      & \$2.4& Recharge base fee - $\rho^r$ & \$1 \\ \hline
Trip cost/mile - $p^m_b$      & \$1  & Recharge cost/mile - $p^{r}$ & \$0.1 \\ \hline
Battery size - $b^{\mbox{\footnotesize size}}$ & 66 kWh (200 miles) & Recharge rate -  $\eta^r$ & 300 miles/hour   \\ \hline
\end{tabular}
\vspace{0.3cm}
\end{center}\vspace{0.1cm}
\caption{Simulation Parameters}
\label{tab:sim_params}
\end{table}

\subsection{Parameterized Myopic Policy}
\begin{figure}[t]
\centering
\vspace{-0.5cm}
\subfigure[]{
\includegraphics[scale = 0.52]{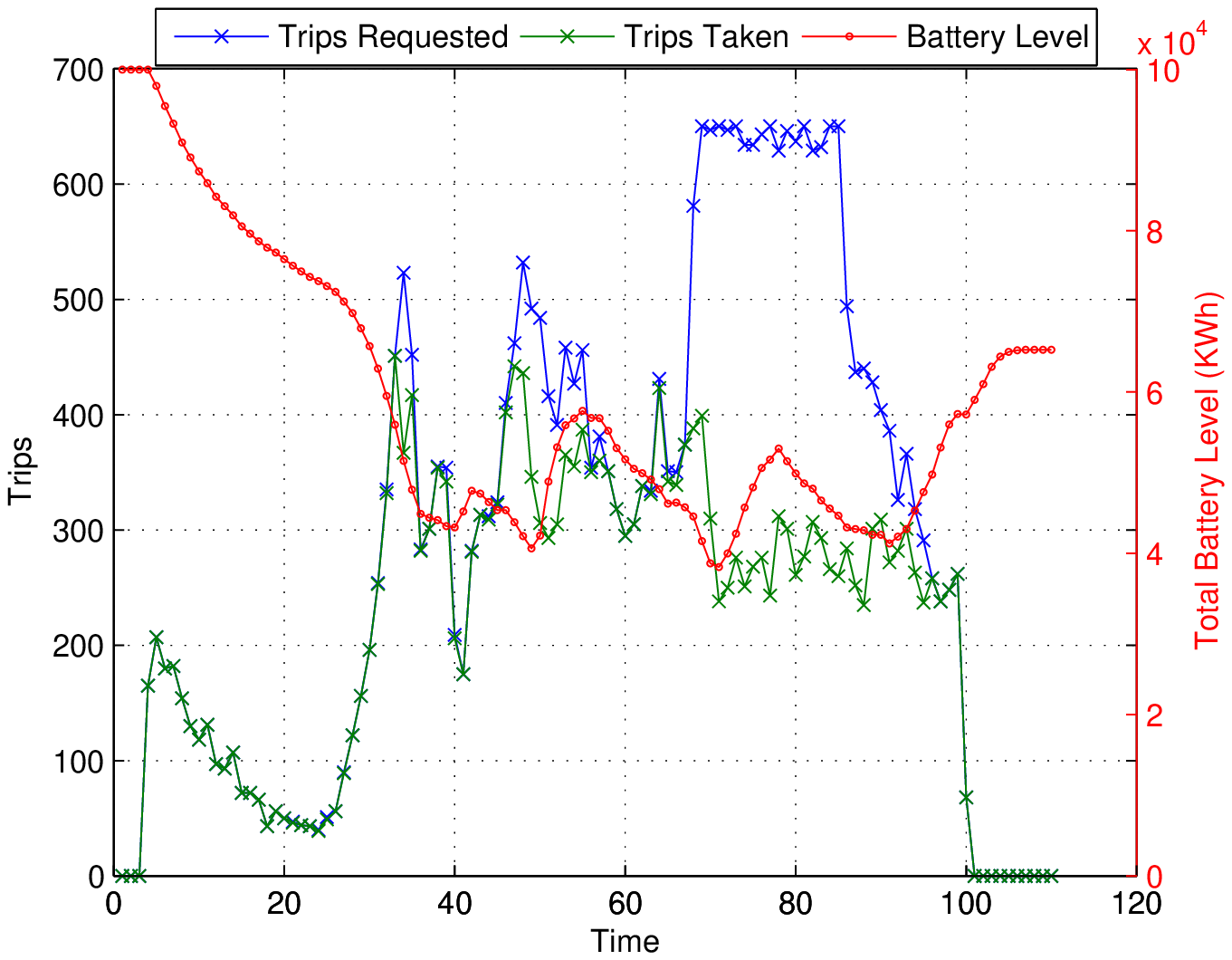}
\vspace{-0.0cm}
\label{fig:Trips_Battery_PFA}
}
\vspace{-0.0cm}
\subfigure[]{
\includegraphics[scale = 0.52]{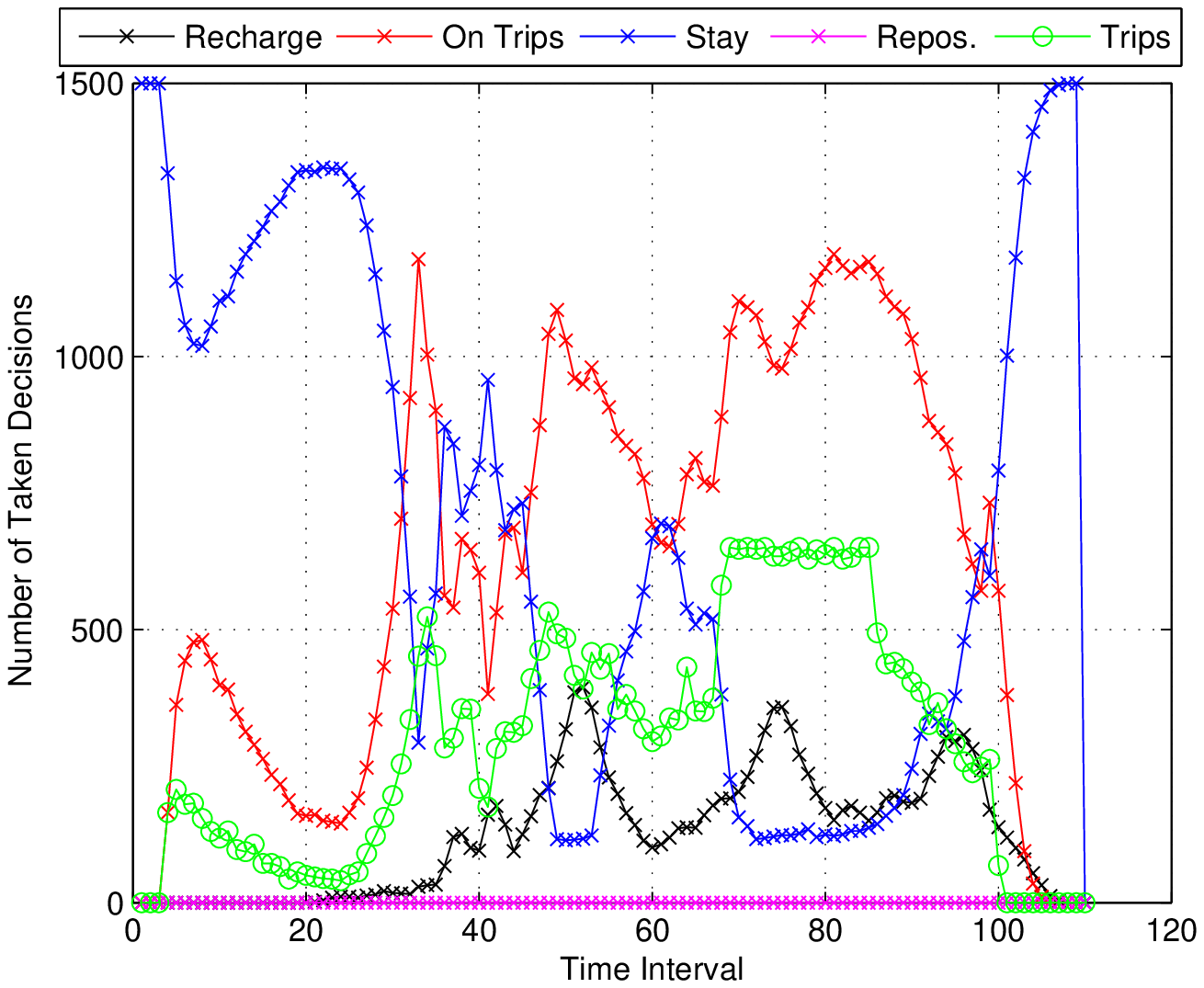}
\vspace{-0.0cm}
\label{fig:Car-Stats-PFA}
}
\caption{\label{fig:Results_Myopic_Policy} Performance results of the myopic policy. (a) Trip coverage and total battery level of the fleet versus time. (b) Number of decisions taken for the cars of the fleet versus time.}
\end{figure}

We begin by studying the performance of the myopic policy given the simulation parameters presented above. This is an online assignment policy that takes milliseconds to be computed at a given time period $t$ as explained in Section~\ref{sec:myopic_policy}. The generated revenue in this case is \$$645,584$ with an average revenue of \$$430$ per car. The performance metrics in terms of trip coverage and car activity statistics are presented in Figure~\ref{fig:Results_Myopic_Policy}.

Figure~\ref{fig:Trips_Battery_PFA} shows the distribution of trip requests, the number of trips taken and the total battery level in the fleet at each point in time. The average trip coverage achieved throughout the day is $73.8$\%. The fleet starts fully charged where every car has a full battery at the beginning of the day, then as time goes on where the cars are taking trips, the battery level decreases in the fleet. Using the  myopic policy, the cars recharge only when the battery level drops below a certain threshold $\theta^{\mbox{thr}}$ which is set to be $10$\% in this case. Figure~\ref{fig:Trips_Battery_PFA} shows that the fleet reaches its lowest battery level just before the peak afternoon period, forcing the cars to recharge. So, we see a significant drop in trip coverage since the cars are recharging during the peak period and they are overcharged by the end of the day which is not needed. 

The behaviour of the battery level in the fleet can be understood by investigating the activities of the cars which are shown in Figure~\ref{fig:Car-Stats-PFA}; it shows the number of cars taking trips (On Trips), staying, repositioning and recharging at each point in time with the following~observations:
\begin{itemize}
\item[1.] The cars do not start recharging until mid day with a significant number of cars recharging during the peak period which decreases the trip coverage.
\item[2.] Another disadvantage is that the cars are not repositioned to places where trips are incoming so there are tens of cars sitting in isolated places during the peak period.
\end{itemize}
Table~\ref{tab:comp_policies} compares the performance results of the various policies that we discussed in Section~\ref{sec:algorithmicstrategies}. The results of VFA with and without hierarchical aggregation (HA) will be discussed below.

\begin{table} [tb]
\begin{center}
\begin{tabular}{|l|c|c|c|} \hline
Metric    & Myopic & VFA without hier. agg. & VFA with hier. agg. \\ \hline \hline
Revenue                &\$645,584 & \$606,975  & \$757,410  \\ \hline
Rvenue/Car             &\$430 & \$404 &  \$505  \\ \hline
Trip Coverage          &73.8\% & 64.9\% & 95\%  \\ \hline
\% Cars on Trips       &48.3\%& 43.77\% & 59\%  \\ \hline
\% Cars Staying        &41.5\% & 40.44\%&  20.1\% \\ \hline
\% Cars Repositioning  & 0\%& 10.76\%& 13.5\%   \\ \hline
\% Cars Recharging     &9.2\% &5.03\% &  7.4\% \\ \hline
\end{tabular}
\vspace{0.3cm}
\end{center}\vspace{-0.0cm}
\caption{Comparison of performance metrics for the various policies.}
\label{tab:comp_policies}
\end{table}

\subsection{VFA Policy}
\begin{figure}[t!]
\centering
\includegraphics[scale=0.5]{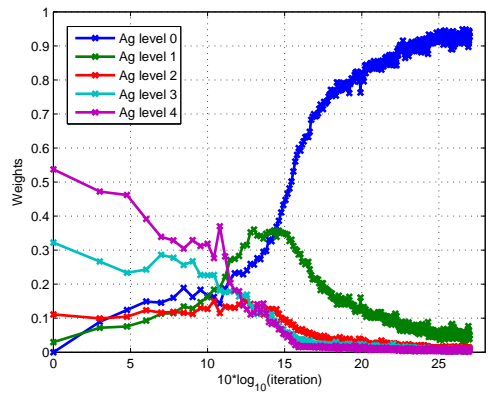}
\caption{\label{fig:aggweights} Weights at different aggregation levels versus the number of iterations in log scale.}
\end{figure}

We demonstrate the effect of capturing the impact of decisions now on the future through value function approximations and the performance gains using hierarchical aggregation. 
Figure \ref{fig:aggweights} shows the average weight put on each level of aggregation from one run of the model.  As is apparent from the figure, higher weights are initially put on the more aggregate estimates, with the weight shifting to the more disaggregate estimates as the algorithm progresses.  

\begin{figure}[t]
\centering
\vspace{-0.5cm}
\subfigure[]{
\includegraphics[scale = 0.52]{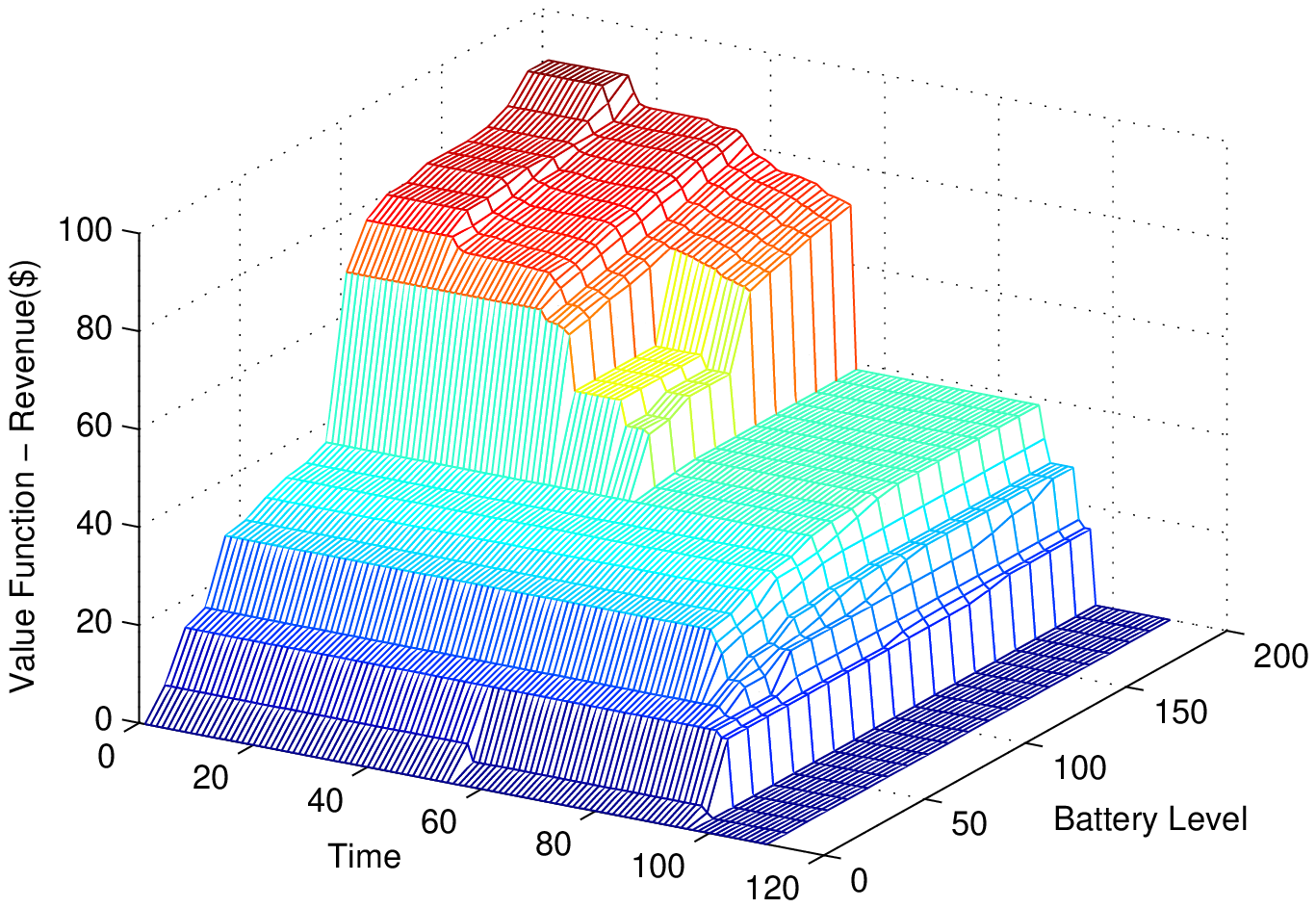}
\vspace{-0.0cm}
\label{fig:VFA_NoHA}
}
\vspace{-0.0cm}
\subfigure[]{
\includegraphics[scale = 0.52]{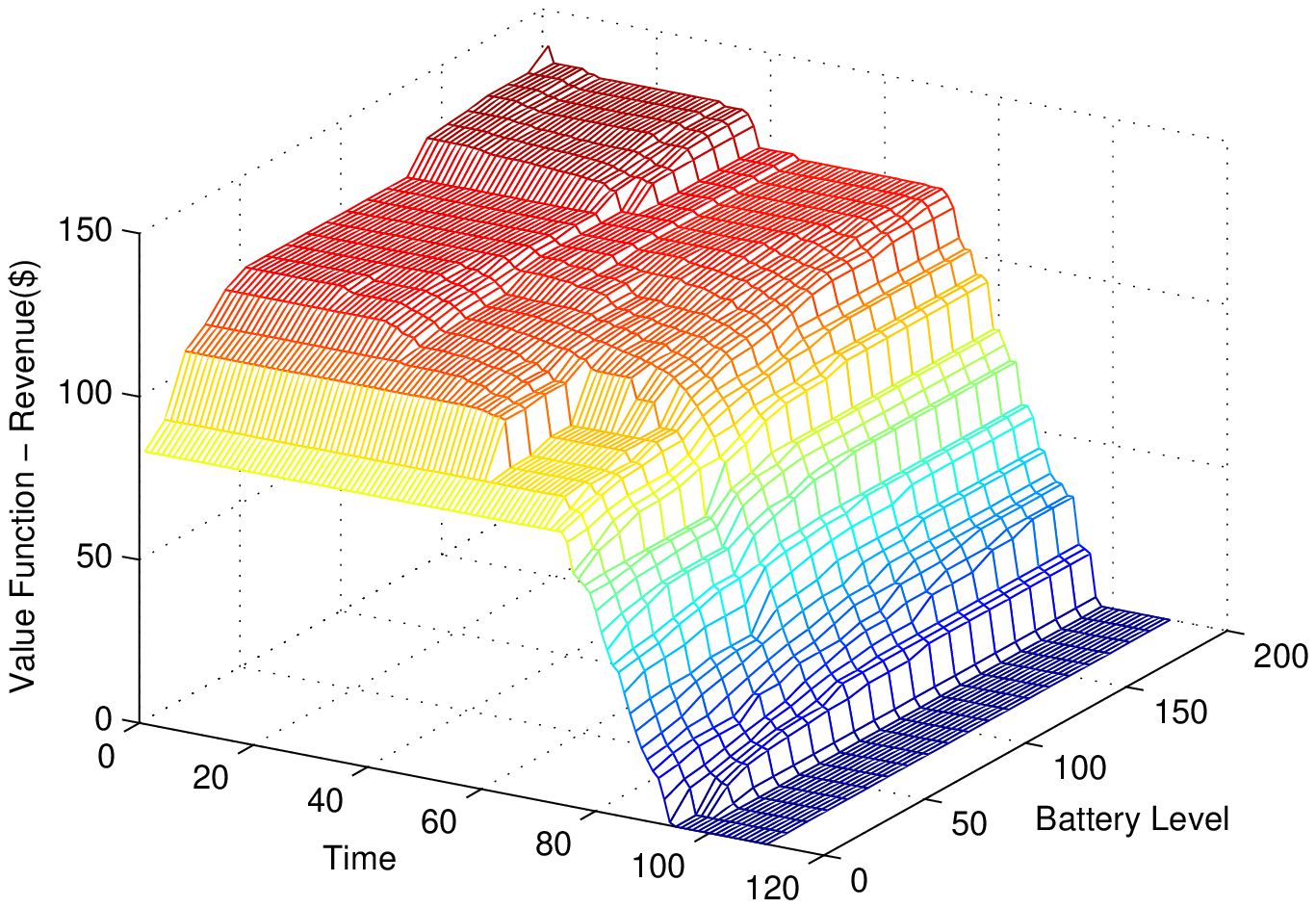}
\vspace{-0.0cm}
\label{fig:VFA_HA}
}
\caption{\label{fig:VFA_W/O_HA} Value function approximation as a function of battery level (miles) for a particular zone and time a) without hierarchical aggregation and b) with hierarchical aggregation. The battery level is presented as a function of how many miles a car can travel before being recharged.}
\end{figure}

Figure~\ref{fig:VFA_W/O_HA} shows the value function of a chosen zone as a function of the time and the battery level with and without hierarchical aggregation.  The value function with hierarchical aggregation is much smoother than the case without hierarchical aggregation because more observations (of cars) including the ones of the neighboring zones were used to evaluate the value function. 

\begin{figure}[t!]
\centering
\includegraphics[scale=0.5]{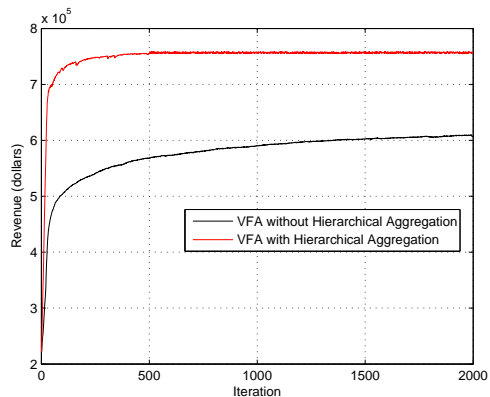}\vspace{-0.2cm}
\caption{\label{fig:Profit} Revenue versus the number iteration for ADP with and without hierarchical aggregation.}
\end{figure}

Figure~\ref{fig:Profit} shows the revenue versus the number of iterations for the proposed approximate dynamic programming algorithm with and without hierarchical aggregation. The results demonstrate that the use of hierarchical aggregation for the VFA produces better results and much faster convergence. 

\begin{figure}[t]
\centering
\vspace{-0.5cm}
\subfigure[]{
\includegraphics[scale = 0.53]{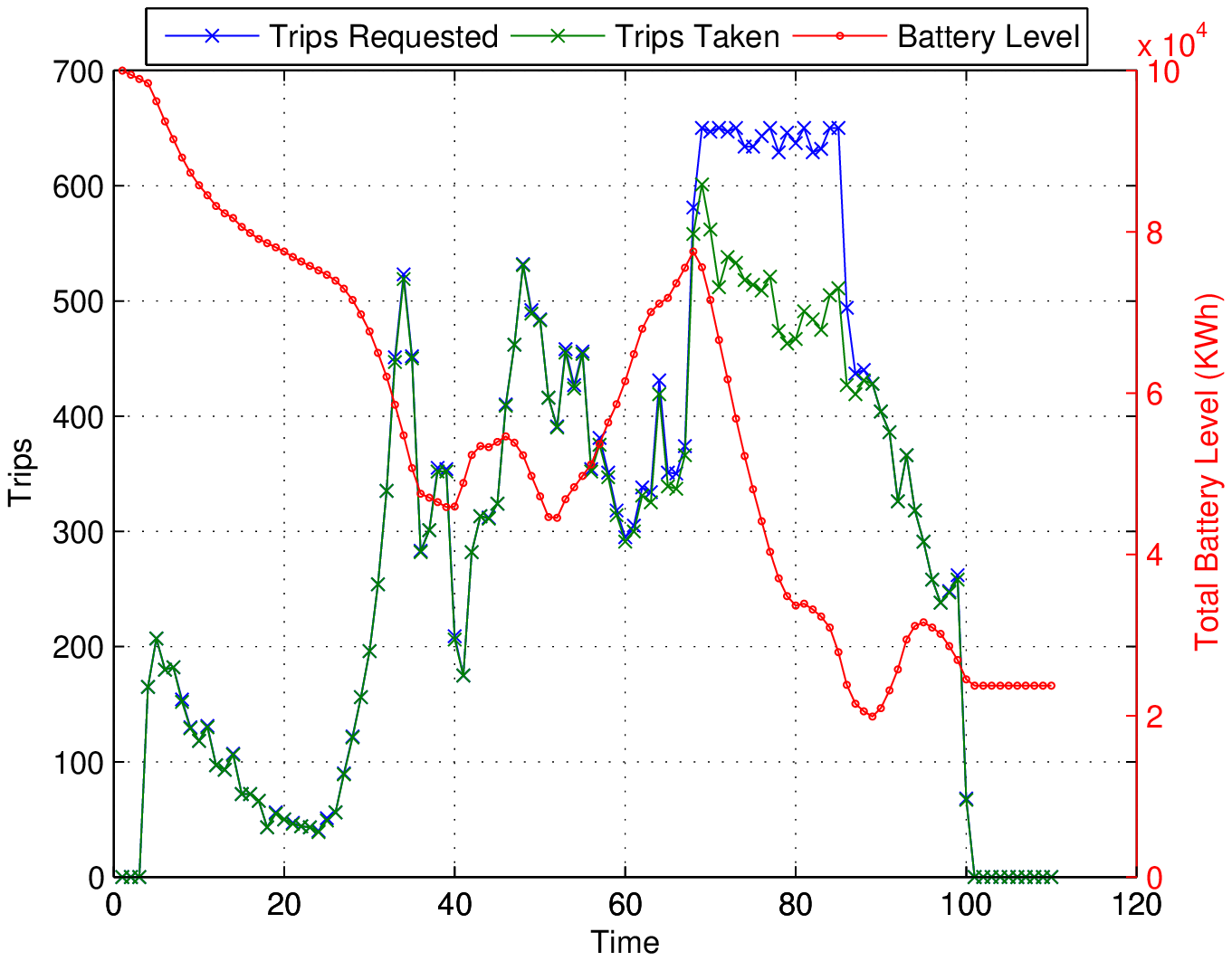}
\vspace{-0.0cm}
\label{fig:Trips_Battery_VFA}
}
\vspace{-0.0cm}
\subfigure[]{
\includegraphics[scale = 0.52]{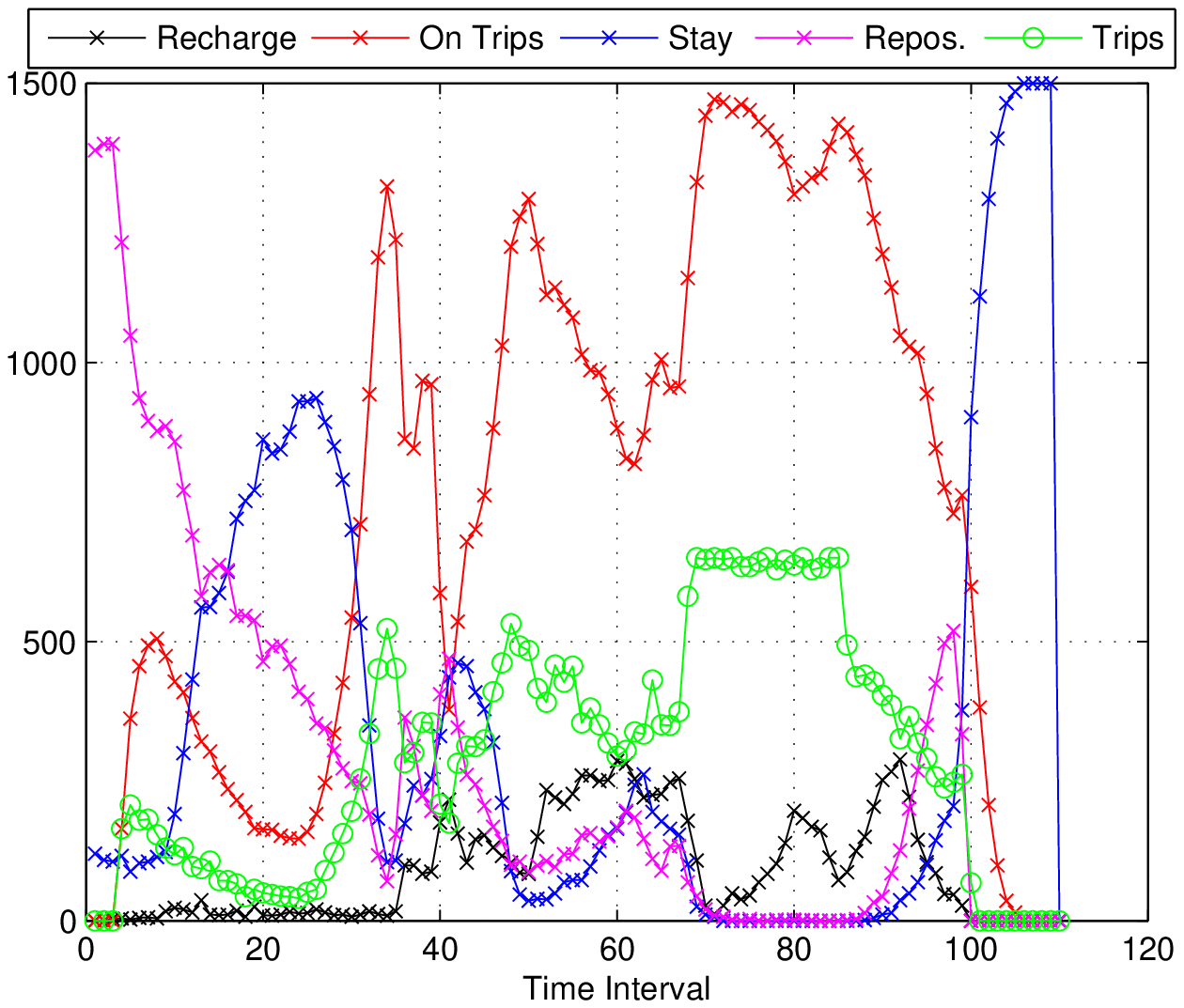}
\vspace{-0.0cm}
\label{fig:Car_Stats_VFA}
}
\caption{\label{fig:Results_VFA_Policy} Performance results of the VFA policy with hierarchical aggregation. (a) Trip coverage and total battery level of the fleet versus time. (b) Number of decisions taken for the cars of the fleet versus~time.}
\end{figure}\vspace{-0.0cm}
Figure~\ref{fig:Trips_Battery_VFA} shows the distribution of trip requests, the number of trips taken and the total battery level in the fleet for the VFA policy with hierarchical aggregation. The average trip coverage achieved throughout the day is $95$\% and the revenue per car is \$$505$ which is $17.32$\% higher than the revenue obtained using the myopic policy. It also shows that the recharging process starts before the peak period where the battery level increases significantly to fulfill the incoming trips unlike the myopic policy. In addition, the policy is able to foresee that no trips are incoming by the end of the day and thus it minimizes the amount of unnecessary energy in the fleet as time goes on.

Figure~\ref{fig:Car_Stats_VFA} shows the activities of the cars at each point in time which can be summarized as
\begin{itemize}
\item[1. ] A significant number of cars are recharging before the peak trip demand.
\item[2. ] In off-peak periods, cars are kept staying or are repositioned.  The VFA policy repositions cars to zones where trips are incoming which guarantees a higher trip coverage. We start with random car locations so the repositioning is mainly done at the beignning of the~day.
\item[3. ] During the peak period, all cars are either taking trips or very few are recharging; however, none is staying in its zone or repositioning. This means that the whole fleet is being used fully with the VFA policy unlike the myopic one.
\end{itemize}
These results show how effective the proposed VFA policy is in repositioning, recharging and allocating trips at each point in time while taking into account future trip demands.

\subsection{Surge Pricing}
\begin{figure}[t!]
\centering
\vspace{-0.5cm}
\subfigure[]{
\includegraphics[scale = 0.52]{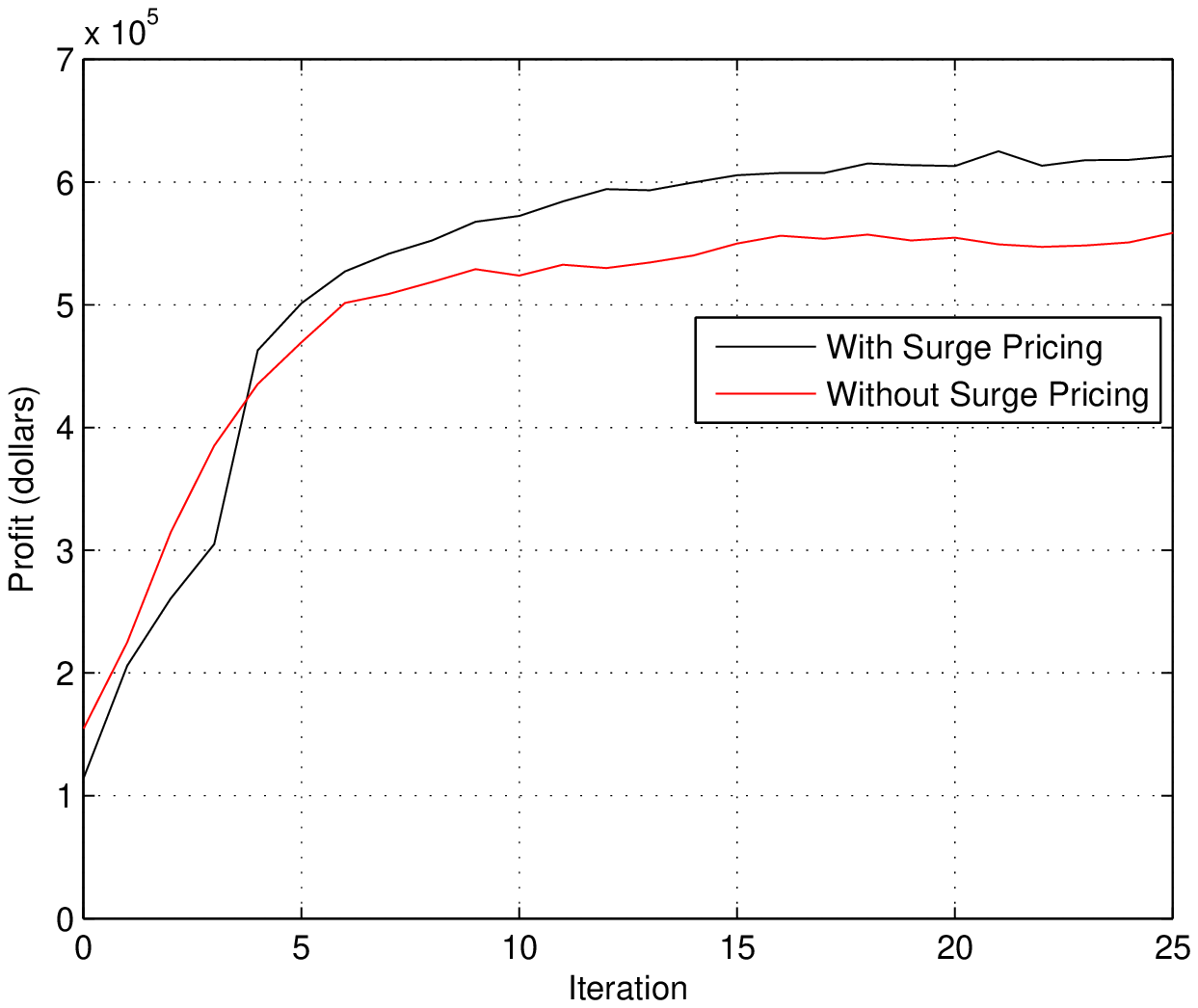}
\vspace{-0.0cm}
\label{fig:profit_500}
}
\vspace{-0.0cm}
\subfigure[]{
\includegraphics[scale = 0.52]{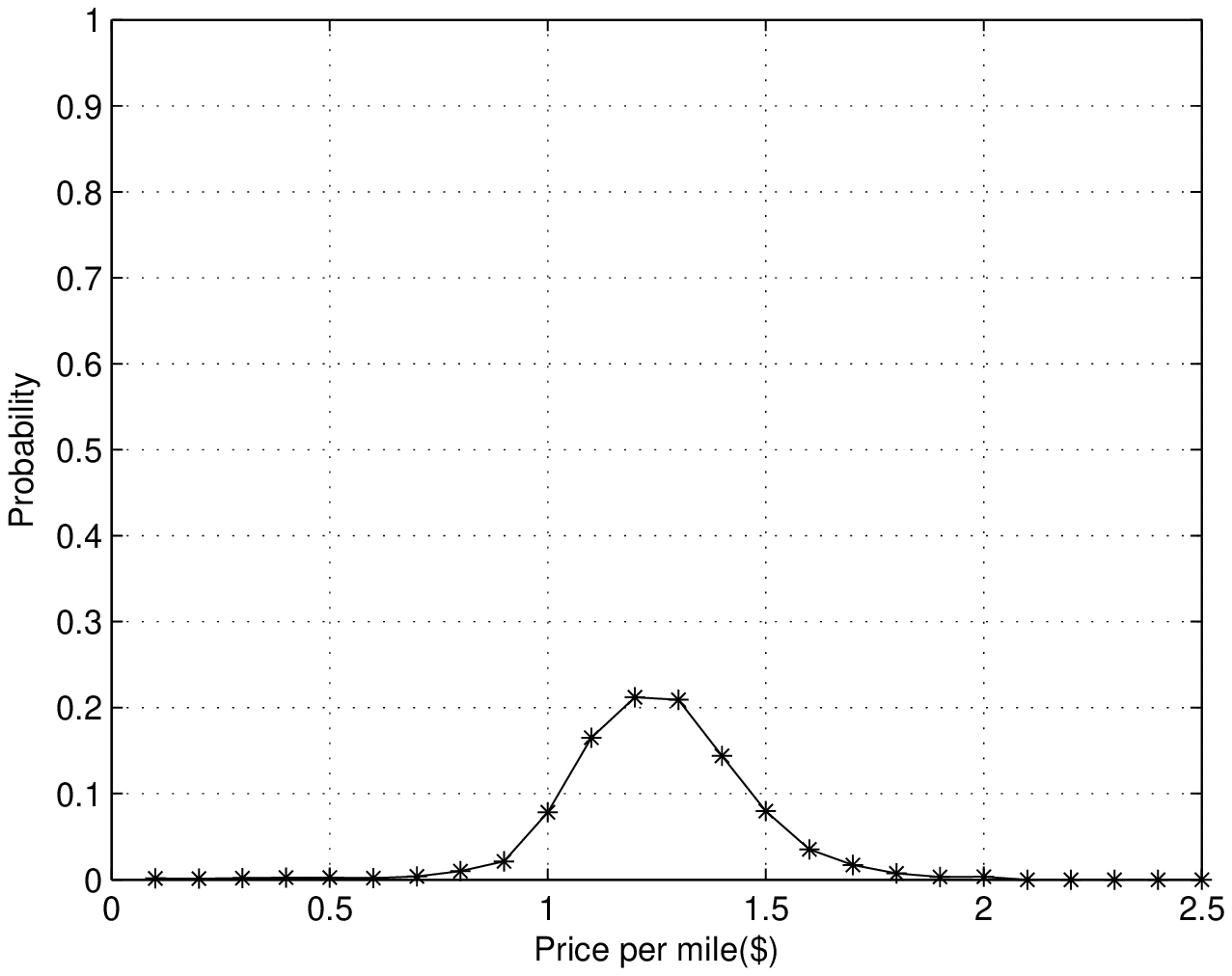}
\vspace{-0.0cm}
\label{fig:price distribution}
}\vspace{-0.0cm}
\caption{\label{fig:profit_surgepricing} Performance results of surge pricing. (a) Revenue with and without surge pricing. (b) Distribution of Prices.}\vspace{-0.5cm}
\end{figure}

For surge pricing, we generate $K=25$ curves for each zone formed of $5$ different curves for each of the operator and rider. Figure~\ref{fig:profit_500} shows the performance gains of surge pricing compared to the case without surge pricing. For the case without surge pricing, we assume that the price per mile is $\$1$. Surge pricing is able to increase the revenue by $13$\%. 

Figure~\ref{fig:price distribution} shows the distribution of prices  with surge pricing over the simulated $24$ hours. It is shown that the price is ranging over an interval between \$$0.5$ and \$$2$ which depends on the minimum and maximum input ranges of the acceptance models and the true response curve of the rider provided initially (see Figure~\ref{fig:acceptance_model} for a one zone example). This indicates that in some places the trip prices are reduced specifically in off-peak periods where the number of available cars is much higher compared to the number of trip requests; however, the prices are increased in peak periods since the trip demand is higher than the number of available~cars and the riders will accept higher prices indicating that the surge pricing model is capturing real riders~behaviour.

\subsection{Economics of the Fleet and Battery Sizes}
\begin{figure}[t!]
\vspace{-0.6cm}
\centering
\includegraphics[scale=0.7]{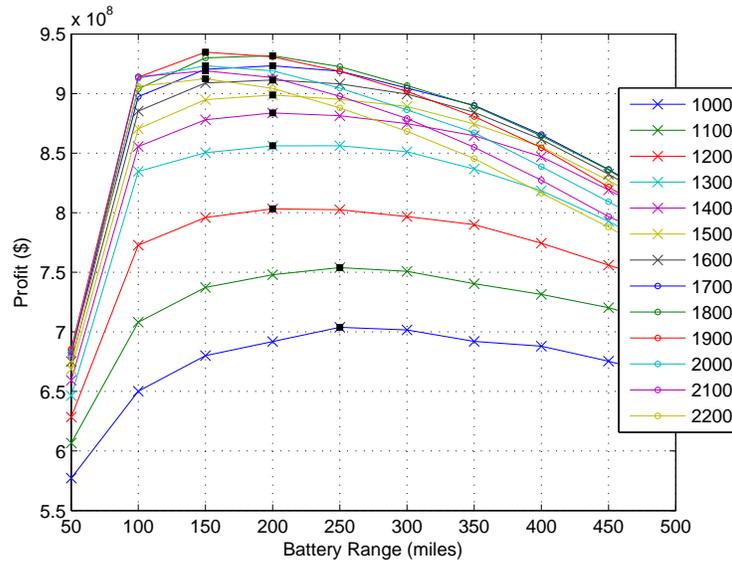}\vspace{-0.2cm}
\caption{\label{fig:economics} Economics of the Fleet; Profit versus time and battery range (every kWh enables the car to travel 3 miles). The maximum value of each curve is represented by a black dot.}
\vspace{-0.2cm}
\end{figure}
We have used the following parameters to compute the economics of the fleet; $y^c=4$ years, $d^o=340$ days per year, $c^{\mbox{\footnotesize car}}=\$40,000$, the maintenance and insurance costs per year, $c^{\mbox{\footnotesize maint}} =\$3000$. As for the capital cost of the battery, $c^{\mbox{\footnotesize bat}}(b^{\mbox{\footnotesize size}})$, we have used the following model. The cost starts with $\$240$/kWh for the first $16.67$ kWhs (which corresponds to a 50 miles range) and then cost increases by 20\% for the next $16.67$ kWhs. Let $b^{\mbox{\footnotesize size}} = 16.67\cdot i$ for $i=\{1, 2,\ldots,10\}$, then the battery cost is $c^{\mbox{\footnotesize bat}}(b^{\mbox{\footnotesize size}})= \$240(1+0.2(i-1))(16.67\cdot i)$. Substituting these parameters in \eqref{eq:economics} and optimizing the revenue for the various fleet and battery sizes gives Figure~\ref{fig:economics}.

Figure~\ref{fig:economics} shows that as the fleet size increases the optimal battery size, that maximizes profit, decreases.
If the fleet size or the battery size is very small, then not all trips can be covered which reduces the profit. Large fleets are also not profitable since many cars are barely needed when their cost is high. Thus, Figure~\ref{fig:economics} shows that medium sized batteries (with a range 150-250 miles) are needed to operate the fleet. For the considered operating costs and trip data set, a good performing fleet size is between 1500 and 1900 cars depending on the used battery size/range and the optimal fleet size is $1900$ cars with a battery range of $150$~miles.

\section{Conclusion}\label{sec:conclusion}
This work modelled a ride sharing system using an autonomous fleet of electric vehicles. Solving the dispatch problem using approximate dynamic programming with the monotinicity of the value functions and hierarchical aggregation produces an accurate simulation of a large-scale fleet. Performance results have shown that the proposed VFA policy outperforms a myopic policy by 17\%. The adaptive learning framework for surge pricing demonstrated an operating behaviour similar to real world where the prices increase with the demand and it has shown a 13\% increase in revenue. It has been also shown that as the fleet size increases the required battery size decreases. Moreover, the logic is able to handle different types of uncertainty including random customer demands and travel times. 

This research will help to quantify the economics of an autonomous fleet. Answering the questions above will help us address other important problems (not addressed in this work)  such as 1) What is the size and distribution of parking lots that would be required to park vehicles that are not needed during off-peak periods? 2) What distribution of charging stations will be needed to serve the vehicles? 3) What is the distribution of battery sizes required to serve the market? 

\begin{APPENDICES}
\section{Proof of Proposition 1}\label{app:proposition1}
To prove that the post-decision value function is monotonically increasing with the battery level, we have also to prove that the pre-deicison value function is monotically increasing in the battery level.

We recall equations  \eqref{eq:bellman2} and \eqref{eq:bellman1} that break the dynamic programming recursions into two steps:
\bns
V^*_t(S_t) &=& \max_{x_t \in \Xcal_t} \big(C(S_t,x_t) + V^{*,x}_t(S^x_t) \big),  \\
V^{*,x}_{t}(S^x_{t}) &=& \E \left\{ V^*_{t+1}(S_{t+1}) \mid S^x_{t} \right\},
\ens
where $S_{t+1} = S^{M,W}(S^x_{t},W_{t+1})$ and $S^x_t = S^{M,x}(S_t,x_t)$.

A state corresponds to $S_t=(R_t,D_t)$ where $R_t = \left\{R_{ta}\right\}_{a\in\mathcal{A}}$ where $R_{ta}$ is the number of cars with attribute~$a$. To prove that the value function of a zone is monotically increasing with the battery level, assume that there is one car in the system, then we need to track how the attribute of this car is changing. In this case, the state of the system $S_t$ is equivalent to $a_t$.  Let $a_t$ be one of the attribute vectors; the first attribute of $a_t$ corresponds to a zone location whereas the second attribute is the battery level. We use $a_t$ instead of $S_t$ as the state and we also have $a_{t+1}=a^{M,W}(a_t,d,W_{t+1})$ and $a^x_{t}=a^{M}(a_t,d)$. For notational consistency, the monotinicity property presented in \eqref{eq:monotone_battery} is equivalent~to:
$$a_t\preceq a_t'\rightarrow V_t(a_t)\leq V_t(a_t').$$

Now, we can prove Proposition 1.

\begin{proof}
We will do the proof by backward induction starting from the base case of $V^*_T$ which satisties \eqref{eq:monotone_battery} by definition. Consider two states, $a_t=(a_{t1},a_{t2})$ and $a_t'=(a_{t1},a_{t2}')$ where $a_t\preceq a_t'$. If the same decicion is applied to both states, then the same battery level is consumed which means that $a^x_t\preceq a_t'^x$. Given $V^*_{t+1}(a_{t+1})$ satisfies \eqref{eq:monotone_battery}, applying (i) and (iii), and the monotinicity of the conditional expectation, we can prove that the post-decision state $V^{*,x}_{t}(a_t^x) = E[V^*_{t+1}(a^{(M,W)}(a_t,d,W_{t+1}))|S_t = a_t, x_t = d]$ is monotone as follows:
\bns
E[V^*_{t+1}(a^{(M,W)}(a_t,d,W_{t+1}))|S_t = a_t, x_t = d] &=& E[V^*_{t+1}(a^{(M,W)}(a_t,d,W_{t+1}))|S_t = (a_{t1},a_{t2}), x_t = d]\\
&\leq& E[V^*_{t+1}(a^{(M,W)}(a_t',d,W_{t+1}))|S_t = (a_{t1},a_{t2}'), x_t = d]\\
&=& E[V^*_{t+1}(a^{(M,W)}(a_t',d,W_{t+1}))|S_t = a_t', x_t = d]
\ens

By this, we have shown that if $V^*_{t+1}(a_{t+1})$ satisfies \eqref{eq:monotone_battery}, then $V^{*,x}_{t}(a_t^x)$ also satisfies \eqref{eq:monotone_battery}, i.e.,
\bns
a^x_t\preceq a_t'^x \rightarrow V^{*,x}_{t}(a_t^x)\leq  V^{*,x}_{t}(a_t'^x)
\ens
Using the second assumption of the proposition, i.e., (ii), we can show that $V^*_{t}(a_t)$ is monotone for any $a_t$:
 \bns
V^*_t(a_t) &=& \arg\max_{x_t\in \mathcal{X}_t}\left(C(a_t,x_t) + V^{*,x}_t(a_t^x)\right) \\
           &\leq& \arg\max_{x_t\in \mathcal{X}'_t}\left(C(a_t',x_t) + V^{*,x}_{t}(a_t'^x)\right)\\
           &=& V^*_{t}(a_t')
\ens
In the inequality, we have $\mathcal{X}_t\subseteq \mathcal{X}'_t$ because we can cover longer trips with a car that has a highher battery level, $a_{t2}'$, compared to a car with a smaller battery level, $a_{t2}$, that cannot cover long trips; hence, the set of feasible deicsions $\mathcal{X}'_t$ is larger than $\mathcal{X}_t$. By this, we complete the proof that the pre-decision and post-decision states satisfy \eqref{eq:monotone_battery}.
\end{proof}

\section{Proof of Proposition 2}\label{app:proposition2}
\begin{proof}
The marginal value of a car with attribute $a=(a_{1},a_2)$ at time $t$ is equivalent to the value (revenue) that it will return if it starts operating in zone $a_1$ with battery $a_2$ from time $t$ up to $T$ without any recharging while operating (this is the value of the car with this specific battery level by definition).  Let $t'\geq t"$, we have:
\bns
v_{t'a} &=& \mathbb{E}\left\{\sum_{t=t'}^T  C_t(S_t,x_t)|S_t = a \right\} \\
       &\geq& \mathbb{E}\left\{\sum_{t=t^{"}}^T  C_t(S_t,x_t)|S_t = a \right\}\\
       &=& v_{t^{"}a}
\ens
This holds since all actions have a reward of $C_t(S_t,x_t)\geq 0$ because no recharging is allowed which is the only case where the car has to pay, i.e., $C_t(S_t,x_t)< 0$. Thus, if we have the same distribution of exogeneous events (trip requests) over the time horizon, then dropping a car earlier in the same zone with the same battery level, will have a higher return; this is due to the fact that the car can collect more revenue (trips) if it operates for a longer time. This concludes the proof of the proposition.
\end{proof}

\setlength{\bibsep}{1.0pt}
\bibliographystyle{elsarticle-harv}
\bibliography{library}
\vspace{0cm}

\end{APPENDICES}
\end{document}